\pdfoutput=1
\documentclass[11pt]{article}
\usepackage[final]{acl}
\usepackage{times}
\usepackage{latexsym}
\usepackage[T1]{fontenc}
\usepackage[utf8]{inputenc}
\usepackage{microtype}
\usepackage{inconsolata}
\usepackage{graphicx}
\usepackage{algorithm}
\usepackage{algorithmic}
\usepackage{newfloat}
\usepackage{listings}
\DeclareCaptionStyle{ruled}{labelfont=normalfont,labelsep=colon,strut=off}
\floatstyle{ruled}
\newfloat{listing}{tb}{lst}{}
\floatname{listing}{Listing}
\usepackage{booktabs}
\usepackage{multirow}
\usepackage{enumitem}
\usepackage{amsmath,amsfonts,amssymb}
\usepackage{makecell}
\usepackage{placeins}
\title{Towards AI-Assisted Research Writing: \\ Benchmarking LLMs for AI/ML Introduction Generation}
\author{
Krishna Garg\textsuperscript{1}\thanks{Equal contribution. Work performed while at University of Illinois Chicago.} \and
Firoz Shaik\textsuperscript{2}\footnotemark[1] \and
Sambaran Bandyopadhyay\textsuperscript{3} \and
Cornelia Caragea\textsuperscript{2} \\[0.5em]
\textsuperscript{1}University of Oxford \quad
\textsuperscript{2}University of Illinois Chicago \quad
\textsuperscript{3}Adobe Research, India \\
{\color{blue}\texttt{krishna.garg@eng.ox.ac.uk \mbox{    }  fshaik8@uic.edu \mbox{    } sambaranb@adobe.com \mbox{    } cornelia@uic.edu}} \\[0.8em]
}
\begin{document}
\maketitle
\begin{abstract}
As researchers increasingly adopt LLMs as writing assistants, generating high-quality research paper introductions remains both challenging and essential.
In this work, we benchmark \textit{Scientific Introduction Generation (SciIG)}, evaluating how well LLMs can produce coherent introductions from titles, abstracts, and related works in the Machine Learning domain. 
Curating new datasets from NAACL 2025 and ICLR 2025 papers, we assess five state-of-the-art models, including both open-source (DeepSeek-v3, Gemma-3-12B, LLaMA 4-Maverick, MistralAI Small 3.1) and closed-source GPT-4o systems across multiple dimensions: lexical overlap, semantic similarity, content coverage, faithfulness, consistency, citation correctness, and narrative quality. Our comprehensive framework combines automated metrics with LLM-as-a-judge evaluations. Results demonstrate LLaMA-4 Maverick's superior performance on most metrics, particularly in lexical overlap, content coverage, citation quality and perplexity. Moreover, three-shot prompting outperforms fewer-shot approaches more than 50\% of the time on formula-based automated metrics and performs consistently better on all human evaluation metrics. These findings provide practical insights into developing effective research writing assistants and set realistic expectations for LLM-assisted academic writing. To foster reproducibility and future research, we publicly release all code and datasets.\footnote{Project page, code, and datasets: \url{https://kgarg8.github.io/emnlp26-sciig/}}
\end{abstract}
\section{Introduction}
The rapid advancement of generative artificial intelligence, particularly large language models (LLMs), like OpenAI's GPT series \citep{brown2020language,achiam2023gpt}, has sparked significant interest in their potential to assist in various complex tasks, including academic writing \citep{jaisankar2025deep}. 
%As researchers and academics strive to balance the increasing demands of scholarly communication with the need for high-quality writing, the importance of using LLMs to generate coherent, engaging, and accurate research paper introductions becomes evident. Despite the impressive capabilities demonstrated by these models, creating an introduction that encapsulates the essence of a research paper, its motivation, scope, and significance remain a formidable challenge. This challenge is amplified by the need for the generated text to be not only contextually relevant but also to exhibit a high degree of narrative quality and academic rigor, as the facts mentioned in a research introduction should be grounded on state-of-the-art literature.
This interest is further driven by the rapid growth of research output in computer science, with recent publication cycles reporting more than 50,000 papers submitted at AAAI 2027, over 17,000 at EMNLP 2026, approximately 19,000 at ICLR 2026, and over 40,000 at NeurIPS 2026. At this scale of scholarly production, maintaining clarity, motivation, and high writing quality places a substantial burden on researchers, making scalable assistance for drafting research paper introductions particularly important. Despite the impressive capabilities of LLMs, it is a great challenge to produce introductions that coherently articulate motivation, scope, and significance while adhering to academic rigor and grounding claims in state-of-the-art literature.

Existing works in the realm of LLM-assisted academic writing have made strides in various domains, from drafting essays to summarizing scientific papers \citep{xiao-etal-2022-primera,nandy2025language, li-ouyang-2025-explaining, wang-etal-2019-paperrobot}. However, these efforts often fall short in several critical aspects. Current models often struggle with maintaining faithfulness to the source material, ensuring consistency in narrative flow, and accurately incorporating citations \citep{huang2025survey}. Moreover, while there have been attempts to measure the quality of generated texts using automated metrics \cite{garg2026large}, these evaluations often overlook nuanced aspects such as the coherence of the introduction and its alignment with academic standards \citep{li2024llms}. There are a plethora of blogs \cite{hlava2025trying, wiggins2024rapid, dataquarry_framework} and videos on quick suggestions on how to use LLMs to write different parts of a research paper, but none of those shows a systematic analysis of the capabilities of current LLMs. There are a few publicly available datasets that maintain a collection of research papers for various generation tasks \citep{lu2020multi,bigsurvey}. However, they are not appropriate for use with the current LLMs since they all have probably appeared in the LLM training set. Consequently, the potential of LLMs in generating high-quality research paper introductions remains underexplored and inadequately benchmarked.

To address these research gaps, we propose a comprehensive framework for the generation of scientific introductions. This framework aims to rigorously evaluate the capability of LLMs to produce introductions of research papers using new datasets curated from the NAACL 2025 and ICLR 2025 accepted papers. Our approach involves evaluating multiple dimensions of generated introductions, including lexical overlap, semantic similarity, content coverage, faithfulness, consistency, citation correctness, and narrative quality. By combining automated metrics with LLM-as-a-judge evaluations \citep{liu2023g}, we seek to provide a holistic understanding of the strengths and limitations of current state-of-the-art models. Finally, a small scale but rigorous human evaluation is conducted to understand the gap between the generated introductions and the expected standard.

The following are the \textbf{contributions} of this paper. 
% First, we emphasize the \textbf{practical task importance} of using LLMs as intelligent assistants for writing research paper introductions: a cognitively demanding component of academic writing that requires synthesizing literature, articulating motivation, and establishing research gaps. 
% Second, we curate two large-scale benchmark \textbf{datasets} comprising 3,900 full-length papers from premier venues, reflecting a realistic input setup of \textit{Title + Abstract + Related Work} that mirrors how researchers approach writing. 
% Third, we develop a \textbf{comprehensive evaluation suite} consisting of 24 metrics that combine traditional NLP measures with custom-designed metrics tailored to this task, such as citation accuracy, consistency, and faithfulness. 
% Fourth, we conduct extensive experiments with five open- and closed-source state-of-the-art LLMs, highlighting the superior performance of LLaMA-4-Maverick, particularly in terms of semantic similarity and faithfulness. 
% Finally, we demonstrate the effectiveness of \textbf{three-shot prompting} over fewer-shot approaches, providing practical guidelines for optimizing LLM-assisted scientific writing workflows. 
% Together, these contributions advance the understanding of LLM capabilities in academic writing and lay the foundation for developing more effective, trustworthy, and human-centered research writing assistants.
First, we present the \textbf{first large-scale benchmark and systematic study} of research paper introduction generation in the LLM era, establishing a common experimental and evaluation foundation for assessing LLM capabilities on this task.
Second, we propose a \textbf{comprehensive evaluation suite comprising 30 metrics}, integrating formula-based automatic metrics, LLM-as-a-judge metrics, and human evaluation metrics to capture both surface-level and higher-order qualities such as faithfulness, consistency, and overall writing quality.
Third, we develop a \textbf{novel end-to-end dataset construction pipeline} that combines PDF text extraction, LLM-assisted citation identification from introductions, and structured retrieval of cited papers' titles and abstracts using the SemanticScholar API. Using this pipeline, we curate two new benchmark datasets totaling 3,900 samples, drawn from papers published at NAACL 2025 and ICLR 2025.
Finally, we systematically explore the impact of \textbf{few-shot prompting strategies} on introduction generation performance and complement our automatic evaluations with a \textbf{human evaluation study} on a subset of samples, providing practical insights into effective LLM-assisted scientific writing.

\section{Related Works}
\textbf{Multi-Document Summarization (MDS)}
A significant body of work in multi-document summarization (MDS) has explored techniques to distill information from multiple related sources into coherent summaries. Models like PRIMERA \cite{xiao-etal-2022-primera} have introduced tailored pre-training strategies that enhance a model's ability to integrate information from concatenated documents using encoder-decoder transformers.
\citet{nandy2025language} have proposed an approach to represent a collection of documents in the form of a hierarchical reference layout tree and generate a summary with attribution from this tree with selected in-context examples.
Meanwhile, graph-based methods \cite{liao-etal-2018-abstract, pasunuru-etal-2021-efficiently, bandyopadhyay2026rsfgllm} leverage structured representations such as Abstract Meaning Representation (AMR) and discourse relations to encode inter-document connections. However, these approaches often require additional linguistic annotations or parsing pipelines, limiting their applicability in general settings.

\textbf{Large Language Models for Scientific Summarization}
General-purpose LLMs such as GPT-3.5 \cite{arshad2023performance}, GPT-4 \cite{achiam2023gpt, ouyang2022training}, PaLM \cite{chowdhery2023palm}, LLaMA \cite{touvron2023llama}, Bloom \cite{workshop2022bloom}, and GLaM \cite{du2022glam} have demonstrated impressive performance across summarization benchmarks. While these models excel in single-document generation tasks, their application to multi-input, constraint-driven generation—such as meta-review synthesis or domain-specific aggregation—remains under-investigated \cite{hase-bansal-2022-models}.

\textbf{LLM-based Academic Text Generation}
A growing line of work builds end-to-end systems that use LLMs to produce extended academic text. Closest to our task, \citet{liebling-etal-2025-towards} frame introduction generation as part of an interactive AI-assisted authoring system, in which an LLM augments a human-written draft with retrieved candidate citations and recommends their inline placement, with evaluation centered on the human--LLM interactive loop over a relatively small curated set. We instead formulate Scientific Introduction Generation as a \emph{standalone} benchmark: given fixed inputs (Title, Abstract, Related Work), models generate a complete Introduction in a single pass, without external retrieval, citation recommendation, or human-in-the-loop scaffolding.
At a different scope, AutoSurvey \citep{wang2024autosurvey} autonomously synthesizes prior work into broad surveys, and \citet{bao-etal-2025-surveygen} introduce SurveyGen, a large dataset of human-written surveys paired with a quality-aware retrieval-augmented framework (QUAL-SG) for evaluating LLM-generated surveys; while related in spirit, full-survey generation prioritizes comprehensive coverage rather than the precise argumentation and contextual positioning required by research paper introductions.

\begin{figure*}[t]
  \centering
  \includegraphics[width=\linewidth]{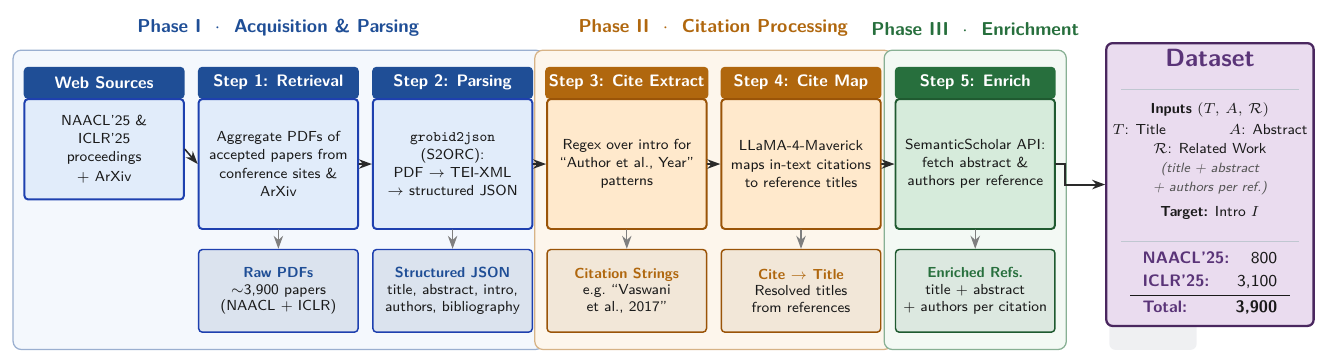}
  \caption{Dataset construction pipeline for the SciIG task. Papers from
  NAACL'25 and ICLR'25 are retrieved and parsed (Phase~I), their citations
  extracted and mapped to reference titles (Phase~II), and each reference
  enriched with Semantic Scholar metadata (Phase~III). Each sample pairs
  $(T, A, \mathcal{R})$ with the published introduction~$I$ as the target.}
  \label{fig:pipeline}
\end{figure*}

\section{Task}
\subsection{Problem Formulation}
The formulation of the task \textit{Scientific Introduction Generation (SciIG)} can be seen as a conditional generation problem. Given a research paper's \textbf{Title} \( \mathcal{T} \), \textbf{Abstract} \( \mathcal{A} \), and a set of \(k\) \textbf{Related Papers} \( \mathcal{R} = \{r_1, r_2, ..., r_k\} \), where each \( r_i \) contains a title, abstract, and author list, the objective is to generate a coherent, academically styled \textbf{Introduction} \( \mathcal{I} \). Formally, the model does a mapping:
\[
f: (\mathcal{T}, \mathcal{A}, \mathcal{R}) \mapsto \mathcal{I}
\]
where $f$ is instantiated by a large language model (LLM) conditioned on a task-specific prompt.

% We choose \textbf{Title}, \textbf{Abstract}, and \textbf{Related Work} as inputs because they collectively provide the essential ingredients for generating effective scientific introductions. According to Swales' widely adopted CARS model \cite{swales1990genre}, introductions typically serve three rhetorical purposes: (1) establishing the research territory, (2) identifying the niche, and (3) occupying that niche. In this context, the \textit{related work} provides a foundation for establishing the territory and highlighting gaps in the literature; the \textit{abstract} conveys the core contribution and its significance, aligning with niche occupation; and the \textit{title} frames the overall research focus. We manually examined 20 NAACL and ICLR papers, all of which followed CARS model.
We choose \textbf{Title}, \textbf{Abstract}, and \textbf{Related Work} as inputs because they provide the minimal yet sufficient information required for generating scientific introductions. Under Swales’ CARS model \cite{swales1990genre}, introductions establish the research territory, identify a gap, and position the contribution. Accordingly, the \textit{related work} supports territory establishment and gap identification, the \textit{abstract} conveys the core contribution, and the \textit{title} frames the research focus. A manual inspection of 20 NAACL and ICLR papers confirms that this structure is consistently followed.

% We deliberately exclude other paper sections to reflect realistic writing workflows and practical benchmarking constraints. Introductions are often drafted \textit{before} later sections are finalized, and conditioning on downstream sections would rely on post-hoc knowledge of results and conclusions that is not available at writing time. Including full paper content would also substantially increase token usage and computational cost, limiting scalability. In addition, reliably extracting and aligning all sections from large-scale PDF collections is challenging due to formatting variability across venues. Restricting inputs to Title, Abstract and Related Work therefore ensures methodological realism, scalability, and reproducible benchmarking.

We deliberately exclude other paper sections from the input to ensure consistent, reproducible benchmarking. Including full paper content would substantially increase token usage and computational cost, limiting scalability to large evaluation sets. In addition, reliably extracting and aligning all sections from large-scale PDF collections is challenging due to formatting variability across venues. %Restricting inputs to Title, Abstract, and Related Work therefore yields a controlled, reproducible benchmark that isolates Introduction-level generative capability and scales across thousands of papers.

\begin{table*}[htbp]
\centering
\small
\renewcommand{\arraystretch}{1.15}
\begin{tabular}{@{}l@{\hspace{1em}}c@{\hspace{1em}}c@{\hspace{1em}}c@{\hspace{1em}}c@{\hspace{1em}}c@{\hspace{1em}}c@{\hspace{1em}}c@{\hspace{1em}}c@{\hspace{1em}}c@{}}
\toprule
& & \multicolumn{3}{c}{\textbf{Introductions (words)}} & \multicolumn{3}{c}{\textbf{Abstracts (words)}} & \multicolumn{2}{c}{\textbf{Citations}} \\
\cmidrule(lr){3-5} \cmidrule(lr){6-8} \cmidrule(lr){9-10}
\textbf{Dataset} & \textbf{Samples} & \textbf{Mean} & \textbf{Median} & \textbf{Range} & \textbf{Mean} & \textbf{Median} & \textbf{Range} & \textbf{Average} & \textbf{Range} \\
\midrule
NAACL 2025 & $800$ & $597.1 \pm 228.4$ & 610.0 & 66--1,546 & $169.3 \pm 76.7$ & 162.0 & 25--1,748 & 13.4 & 1--45 \\
ICLR 2025 & $3100$ & $771.1 \pm 414.3$ & 747.5 & 42--13,675 & $210.4 \pm 101.4$ & 194.0 & 6--1,063 & 15.3 & 1--58 \\
\bottomrule
\end{tabular}
\caption{Statistics for the NAACL 2025 (800 samples) and ICLR 2025 (3100 samples) datasets, including number of samples, text length, and citation metrics, highlighting variability in introduction lengths and citation patterns relevant to the Introduction Generation task.}
\label{tab:dataset-stats}
\end{table*}

\subsection{Datasets}
\label{sec:datasets}
To support the SciIG task, we construct two datasets\footnote{We make the datasets publicly available under CC BY 4.0.} from accepted papers of the NAACL 2025 and ICLR 2025 conferences, comprising 800 and 3100 samples, respectively. 
Our ICLR and NAACL datasets comprise 3,900 high-quality long-form NLP and ML papers, comparable to existing benchmarks \cite{costa-jussa-etal-2025-lcfo, chandrasekaran2019overview, mahata2022ldkp, cachola-etal-2020-tldr}.

The creation process involved five key steps (illustrated in Figure~\ref{fig:pipeline}; full details in Appendix~\ref{app:pipeline}) to extract titles, abstracts, introductions, authors, and citation details, ensuring rich data for the task. At a high level, the pipeline (i) retrieves PDFs of accepted NAACL 2025 and ICLR 2025 papers from official proceedings and ArXiv; (ii) parses them into structured JSON via the S2ORC \texttt{grobid2json} tool \citep{lo-etal-2020-s2orc}; (iii) extracts in-text Related Works citations using a regular expression over ``Author et al., Year'' patterns; (iv) maps these citation strings to their bibliographic entries using the \textbf{LLaMA 4-Maverick} model (prompt in Table~\ref{tab:citation-mapping-prompt}); and (v) enriches each cited title with an abstract and author details via the Semantic Scholar API\footnote{\url{https://www.semanticscholar.org/}} \cite{Kinney2023TheSS}, using both the official and an unofficial\footnote{\url{https://github.com/danielnsilva/semanticscholar}} endpoint. We restrict the corpus to 2025 NAACL and ICLR papers to ensure community-standard editorial quality, preserve domain consistency within ML/NLP, and post-date all evaluated models' pretraining cutoffs. The resulting corpus comprises 3,900 long-form samples with a 94.0\% citation resolution rate. Table~\ref{tab:dataset-stats} summarizes text length and citation statistics, highlighting the datasets' variability and suitability. We document the full multi-stage Semantic Scholar resolution strategy, the handling of partial \texttt{None} responses from the API, and the resulting per-dataset coverage statistics in Appendix~\ref{app:s2resolution}.

The resulting datasets exhibit significant diversity, as shown in Table~\ref{tab:dataset-stats}. NAACL 2025 introductions average $597.1 \pm 228.4$ words, while ICLR 2025 introductions are longer at $771.1 \pm 414.3$ words, with an outlier maximum of 13{,}675 words, posing challenges for generation. Abstracts are similarly variable, with median lengths of 162 words (NAACL) and 194 words (ICLR); a small tail of very short abstracts (fewer than 60 papers, $<2\%$ of the 3{,}900 samples) arises as Grobid parsing artefacts rather than a systematic data issue. Citation patterns within the Introductions also vary, with NAACL 2025 averaging 13.4 citations per paper and ICLR 2025 averaging 15.3, reflecting dense scholarly connections. These characteristics make the datasets ideal for evaluating the Scientific Introduction Generation task across varied text lengths and citation contexts. In our experiments, we use models with sufficient context length to process the entire metadata, accommodating the datasets' variable introduction lengths and citation-rich content.

We limit this initial effort to the ML domain since recruiting domain-expert annotators is challenging. As ML researchers, we annotated the outputs ourselves. However, since no model was trained on this data, and the study focuses purely on benchmarking, the risk of author bias is minimal.

\section{Methods and Evaluation}
\subsection{Methods}
To evaluate the Scientific Introduction Generation task on the NAACL 2025 and ICLR 2025 datasets, we employ five state-of-the-art language models with knowledge cutoffs before January 2025\footnote{Note that the cutoff date of LLM pretraining reduces the probability of overlap but cannot eliminate it entirely since the pretraining data is often unpublished.}: \textbf{Deepseek-V3-0324} \cite{liu2024deepseek}, \textbf{LLaMA 4-Maverick}\footnote{https://huggingface.co/meta-llama/Llama-4-Maverick-17B-128E-Instruct}, \textbf{Mistral-Small-3.1-24b}\footnote{https://huggingface.co/mistralai/Mistral-Small-3.1-24B-Instruct-2503}, \textbf{Gemma-3-12b-it} \cite{team2025gemma}, and \textbf{GPT-4o} \cite{hurst2024gpt}. These models, selected for their diverse architectures and robust text generation capabilities, include a combination of open-source and commercialized systems. %They are well-suited to handle the datasets' variable introduction lengths (up to $13{,}675$ words) and citation-rich metadata. 
Please refer Appendix \ref{sec:model choices} for the detailed justification of model choices.
% Each model has sufficient context length to process the entire input, including titles, abstracts, and related paper details, ensuring comprehensive generation.

We design seven prompting strategies to guide the models in generating introductions, tailored to the task's requirements and the datasets' characteristics. These strategies, including \textit{Short}, \textit{Medium}, \textit{Elaborate}, \textit{One-Shot}, \textit{Two-Shot}, \textit{Three-Shot}, and \textit{AutoCoT}, vary in complexity and context, incorporating the target paper's title, abstract, and JSON-formatted related papers with APA citations. The Short prompt provides minimal context for concise generation, while Medium emphasizes research gaps and citation integration. Elaborate enforces a strict four-paragraph structure (each 100--150 words), detailing context, gaps, contributions, and impact. One-Shot, Two-Shot, and Three-Shot incorporate 1--3 example introductions from a held-out set\footnote{Held-out set is created using additional $54$ samples for NAACL 2025 and $52$ samples for ICLR 2025. It is used to select random examples provided in the prompt for few-shot experiments.}, enabling few-shot learning. AutoCoT uses iterative, self-refining prompts to enhance reasoning for complex scenarios. AutoCoT implementation details are discussed in section Appendix \ref{app:autocot}. All prompts maintain a formal academic tone, enforce APA citation accuracy, and leverage the datasets' citation-rich metadata, ensuring robust and contextually relevant introductions. The prompts for each strategy are discussed in detail in Appendix in Tables~\ref{tab:prompts-1}, \ref{tab:prompts-2}, and \ref{tab:prompts-3}.

% \subsection{Evaluation}
% The evaluation of the Scientific Introduction Generation task employs a comprehensive set of metrics organized into seven key categories: Lexical Overlap, Semantic Similarity, Content Coverage, Faithfulness, Consistency, Citation Correctness and Narrative Quality, as detailed in Table~\ref{tab:results-main}. 

\subsection{Evaluation}
\label{sec:eval}
The evaluation of the Scientific Introduction Generation task employs metrics in seven categories: Lexical Overlap, Semantic Similarity, Content Coverage, Faithfulness, Consistency, Citation Correctness, and Narrative Quality, with results reported in Table~\ref{tab:results-main}. Expanded per-metric definitions, formulae, their derivations and individual descriptions of each metric, and rubric-level details for the LLM-as-a-Judge scores are provided in Appendix~\ref{app:eval-details}.
% The evaluation of the Scientific Introduction Generation task employs metrics in seven categories: Lexical Overlap, Semantic Similarity, Content Coverage, Faithfulness, Consistency, Citation Correctness, and Narrative Quality, with results reported in Table~\ref{tab:results-main}.

\textbf{Lexical Overlap} captures surface-level similarity to the ground-truth introduction via ROUGE-1, ROUGE-2, ROUGE-L \citep{lin-2004-rouge}, BLEU \citep{papineni2002bleu}, and METEOR \citep{banerjee-lavie-2005-meteor}\footnote{Implementations sourced from \url{https://github.com/huggingface/evaluate}}. We include ROUGE primarily for comparability with prior summarization and generation literature, rather than as a sole quality proxy.

\begin{table*}[htbp]
\scriptsize
\centering
\renewcommand{\arraystretch}{0.85}
\resizebox{\textwidth}{!}{
\begin{tabular}{@{}llcccccccccc@{}}
\toprule
\multirow{2}{*}{\textbf{Category}} & \multirow{2}{*}{\textbf{Metric}}
& \multicolumn{5}{c}{\textbf{NAACL 2025}} & \multicolumn{5}{c}{\textbf{ICLR 2025}} \\
\cmidrule(lr){3-7} \cmidrule(lr){8-12}
& & \textbf{Deepseek} & \textbf{Gemma} & \textbf{LLaMA4} & \textbf{Mistral} & \textbf{GPT-4o} & \textbf{Deepseek} & \textbf{Gemma} & \textbf{LLaMA4} & \textbf{Mistral} & \textbf{GPT-4o} \\
\midrule
\multirow{6}{*}{Lexical Overlap}
& ROUGE-1 & 0.4161 & 0.4361 & \textbf{0.4402} & 0.4388 & 0.4282 & 0.3863 & \textbf{0.4342} & 0.4298 & 0.4326 & 0.4154 \\
& ROUGE-2 & 0.1232 & 0.1304 & \textbf{0.1509} & 0.1340 & 0.1310 & 0.1195 & 0.1340 & \textbf{0.1537} & 0.1344 & 0.1289 \\
& ROUGE-L & 0.1649 & 0.1725 & \textbf{0.1894} & 0.1765 & 0.1754 & 0.1531 & 0.1700 & \textbf{0.1830} & 0.1715 & 0.1689 \\
& BLEU    & 0.1175 & 0.1232 & 0.1198 & \textbf{0.1278} & 0.1043 & 0.0781 & \textbf{0.0979} & 0.0874 & 0.0977 & 0.0762 \\
& METEOR  & 0.2669 & 0.2851 & 0.2700 & \textbf{0.2900} & 0.2658 & 0.2245 & \textbf{0.2600} & 0.2418 & 0.2592 & 0.2358 \\
\midrule
\multirow{4}{*}{Semantic Similarity}
& BERTScore             & 0.8415 & \textbf{0.8475} & 0.8448 & 0.8467 & 0.8422 & 0.8377 & \textbf{0.8449} & 0.8403 & 0.8427 & 0.8377 \\
& BLEURT                & 0.3200 & 0.3088 & \textbf{0.3339} & 0.3236 & 0.3262 & 0.3171 & 0.3032 & \textbf{0.3290} & 0.3194 & 0.3225 \\
& Contextual Relevance  & 0.9691 & \textbf{0.9713} & 0.9684 & 0.9694 & 0.9696 & 0.9687 & \textbf{0.9729} & 0.9696 & 0.9700 & 0.9707 \\
\midrule
\multirow{3}{*}{Content Coverage}
& Reference-based Coverage   & 0.4435 & 0.4513 & \textbf{0.4845} & 0.4692 & 0.4713 & 0.4461 & 0.4732 & \textbf{0.4794} & 0.4725 & 0.4697 \\
& Reference-free Coverage    & 0.3913 & 0.3666 & \textbf{0.4367} & 0.3959 & 0.4035 & 0.3547 & 0.3448 & \textbf{0.4097} & 0.3847 & 0.3664 \\
& \textit{LLM-as-a-Judge} & 0.6907 & 0.7051 & 0.7019 & 0.7008 & \textbf{0.7054} & 0.6567 & \textbf{0.6970} & 0.6809 & 0.6833 & 0.6879 \\
\midrule
\multirow{2}{*}{Faithfulness}
% & QA-based Faithfulness        & 0.7794 & 0.7794 & 0.7794 & \textbf{0.7846} & 0.7762 & 0.7761 & \textbf{0.7814} & 0.7772 & 0.7790 & 0.7751 \\
% & Entailment-based Faithfulness & -0.0364 & 0.1461 & 0.1077 & \textbf{0.1533} & 0.0937 & -0.0276 & 0.1647 & 0.1280 & \textbf{0.1660} & 0.1095 \\
& Keyphrase-based Faithfulness  & 0.3138	& 0.2933 & \textbf{0.3455} & 0.3230 & 0.3263 & 0.2892 & 0.2786 & \textbf{0.3318} &	0.3067 & 0.2941 \\
& \textit{LLM-as-a-Judge} & 0.7873 & 0.8042 & \textbf{0.8157} & 0.7874 & 0.8141 & 0.7825 & 0.8088 & \textbf{0.8149} & 0.7861 & 0.8138 \\
\midrule
\multirow{2}{*}{Consistency}
% & QA-based Consistency        & 0.7820 & \textbf{0.7876} & 0.7860 & 0.7861 & 0.7873 & 0.7810 & \textbf{0.7891} & 0.7846 & 0.7842 & 0.7847 \\
% & Entailment-based Consistency & -0.0151 & \textbf{0.1339} & 0.1077 & 0.1319 & 0.0774 & -0.0552 & 0.0923 & 0.0758 & \textbf{0.1015} & 0.0465 \\
& Keyphrase-based Consistency  & 0.2133 & 0.2170 & \textbf{0.2393} & 0.2230 & 0.2248 & 0.2121 & 0.2197 & 0.2214 & 0.2187 & \textbf{0.2208} \\
& \textit{LLM-as-a-Judge} & 0.8788 & 0.8910 & \textbf{0.8954} & 0.8850 & 0.8945 & 0.8583 & \textbf{0.8873} & 0.8826 & 0.8761 & 0.8868 \\
\midrule
\multirow{4}{*}{Citation Correctness}
& Recall              & 0.5158 & \textbf{0.5763} & 0.5574 & 0.4914 & 0.4826 & 0.4954 & 0.5413 & \textbf{0.5505} & 0.4671 & 0.4657 \\
& Precision           & 0.8819 & 0.9049 & \textbf{0.9357} & 0.8889 & 0.9097 & 0.8686 & 0.8965 & \textbf{0.9238} & 0.8822 & 0.8942 \\
& \textit{LLM-as-a-Judge (Cit. Qual.)}    & 0.7683 & 0.7959 & \textbf{0.8051} & 0.7658 & 0.8050 & 0.7575 & \textbf{0.8021} & 0.8019 & 0.7597 & 0.7968 \\
\midrule
\multirow{2}{*}{Narrative Quality}
& Perplexity  $\downarrow$        & 28.5031 & 32.3897 & \textbf{18.4870} & 25.6458 & 21.8494 & 36.9546 & 29.1004 & \textbf{21.1194} & 29.2585 & 22.0046 \\
& \textit{LLM-as-a-Judge} & 0.8892 & 0.8888 & 0.9001 & 0.8928 & \textbf{0.9004} & 0.8835 & 0.8899 & 0.8986 & 0.8907 & \textbf{0.9006} \\
\bottomrule
\end{tabular}
}
\caption{Evaluation metrics categorized across standard and LLM-as-a-Judge criteria for NAACL 2025 and ICLR 2025 using Elaborate prompting strategy. Highest values are bolded. Note: Lower perplexity values indicate better performance.}
\label{tab:results-main}
\end{table*}

\textbf{Semantic Similarity} assesses meaning-level alignment using BERTScore \cite{zhang2019bertscore}, BLEURT \cite{sellam-etal-2020-bleurt}\footnote{Implementation sourced from \url{https://github.com/lucadiliello/bleurt-pytorch}}, and our proposed \textsc{Contextual Relevance}. Let $g \in \mathbb{R}^d$ be the embedding of the generated introduction and $\{c_1, \dots, c_n\}$ the embeddings of its title, abstract, and cited papers. Contextual Relevance is
\[
\frac{1}{n} \sum_{i=1}^n \frac{g \cdot c_i}{|g|\,|c_i| + \epsilon},
\]
with $\epsilon$ a small numerical-stability constant. See Appendix~\ref{app:ctxrel} for implementation details.

\textbf{Content Coverage} measures how much key content the generated introduction captures, either against the ground-truth introduction\footnote{We use \textit{ground truth introduction} for the published introduction, but emphasize it is not a unique gold standard: the same paper admits many valid renderings. RBC therefore measures coverage against one accepted reference, complemented by an RFC counterpart over Title and Abstract keyphrases.} or against the input context (Title and Abstract). Let $\mathcal{K}_{\text{gen}}, \mathcal{K}_{\text{GT}}, \mathcal{K}_{\text{ctx}}$ be the keyphrase sets\footnote{See Appendix~\ref{app:kbert} for keyphrase extraction details.} extracted from the generated introduction, the ground truth, and the Title+Abstract respectively, where two keyphrases satisfy the exact match $k = k'$ only if their stemmed forms are identical.
%with semantic match $k \sim k' \iff \cos(\mathbf{e}(k), \mathbf{e}(k')) \geq \tau$ (e.g., $\tau{=}0.7$). 
\textbf{Reference-based Coverage} is then
\[
\text{RBC} = \frac{|\{k \in \mathcal{K}_{\text{GT}} \mid \exists\, k' \in \mathcal{K}_{\text{gen}},\, k = k'\}|}{|\mathcal{K}_{\text{GT}}|},
\]
% and \textbf{Reference-free Coverage} (RFC) is defined analogously over $\mathcal{K}_{\text{ctx}}$.
\textbf{Reference-free Coverage} (RFC) is defined analogously over the input-context keyphrase set $\mathcal{K}_{\text{ctx}}$:
\[
\text{RFC} = \frac{|\{k \in \mathcal{K}_{\text{ctx}} \mid \exists \, k' \in \mathcal{K}_{\text{gen}},\, k = k'\}|}{|\mathcal{K}_{\text{ctx}}|}.
\]
RBC and RFC share the same keyphrase extractor and the two are complementary. RBC measures recovery of the published introduction's key content, while RFC measures grounding in the explicit input context.

\textbf{Faithfulness} assesses factual alignment between the generated Introduction and its source inputs via a \textsc{Keyphrase-based Faithfulness} metric, which captures the proportion of generated keyphrases that are factually grounded in the source context (Title and Abstract), together with an LLM-as-a-Judge score (Appendix~\ref{sec:appendix-faithfulness}). \textbf{Consistency} mirrors this implementation but uses the ground-truth introduction as the reference instead of the Title and Abstract, assessing whether the generated text maintains logical and factual consistency with the published introduction.

\textbf{Citation Correctness} comprises three metrics. We extract citation strings (e.g., ``et al.'' identifiers) from the generated introduction using LLaMA 4-Maverick (prompt in Table~\ref{tab:citation-mapping-prompt}) and match them against the original Related Works citation list. \textbf{Citation Precision} is the fraction of generated citations that appear in the original list, and \textbf{Citation Recall} is the fraction of original Related Works citations recovered in the generated introduction. An \textbf{LLM-as-a-Judge} score (Citation Contextual Quality) additionally assesses whether the cited sentences accurately reflect the cited paper's metadata and are placed in an appropriate rhetorical context.

\textbf{Narrative Quality} is evaluated using Perplexity and an LLM-as-a-Judge score. Perplexity, computed as the exponential of the average negative log-likelihood under GPT-2 \citep{radford2019language}, indicates fluency, with lower values reflecting more natural text. The judge score complements perplexity by grading grammar, logical flow, academic tone, and readability.

The \textbf{LLM-as-a-Judge} metrics \citep{liu2023g} use GPT-4o to score generations on Content Coverage, Faithfulness, Consistency, Citation Contextual Quality, and Narrative Quality (prompts in Table~\ref{tab:llm-judge-prompts}), complementing the automated metrics with human-like judgment.

\section{Results and Analysis}
\label{sec:results}

% Table~\ref{tab:results-main} presents the performance of five state-of-the-art models (\textbf{Deepseek}, \textbf{Gemma}, \textbf{LLaMA4}, \textbf{Mistral}, \textbf{GPT-4o}) on the \textbf{NAACL 2025} (800 samples) and \textbf{ICLR 2025} (3100 samples) datasets for the SciIG task. The results highlight several important observations.

% First, LLaMA4 achieves consistently high lexical overlap and semantic similarity scores, particularly evident in its superior performance on both datasets. This indicates LLaMA4's strong capability in closely replicating the original reference content, likely due to optimized pretraining and instruction-following mechanisms.

% Second, although Gemma demonstrates exceptional recall in Citation Correctness, it shows relatively lower precision compared to LLaMA4, suggesting effective reference incorporation but occasional introduction of extraneous citations.

% Third, Mistral shows remarkable entailment-based consistency scores, surpassing other models on both datasets, indicating its strength in maintaining logical alignment between input contexts and generated text. However, its higher perplexity suggests a trade-off with fluency.

% Fourth, Gemma's superior Contextual Relevance scores, especially on the ICLR dataset, imply that it captures deeper contextual nuances within research topics. Lastly, despite GPT-4o's competitive QA-based consistency scores, its lower lexical overlap and citation recall metrics suggest challenges in precisely capturing explicit references.

Table~\ref{tab:results-main} presents the performance of five state-of-the-art models (\textbf{DeepSeek}, \textbf{Gemma}, \textbf{LLaMA4-Maverick}, \textbf{Mistral}, and \textbf{GPT-4o}) on the \textbf{NAACL 2025} (800 samples) and \textbf{ICLR 2025} (3100 samples) datasets for the SciIG task. The results yield several key observations.

First, \textbf{LLaMA4-Maverick} emerges as the overall front-runner, achieving superior scores across lexical overlap, content coverage, citation quality, and perplexity metrics, underscoring its balanced capability in producing coherent, accurate, and fluent introductions. 

Second, \textbf{Gemma} performs particularly well on metrics of semantic alignment, including BERTScore and contextual relevance, suggesting its strength in preserving meaning and capturing nuanced topic relationships. 

Third, both \textbf{Gemma} and \textbf{Mistral} demonstrate competitive results on factual grounding metrics such as faithfulness and consistency, indicating their reliability in maintaining factual integrity across generated texts. 

Finally, \textbf{GPT-4o} and \textbf{DeepSeek} trail behind on most metrics, reflecting challenges in achieving consistent factual accuracy and precise lexical alignment for this specialized academic writing task.

The \textbf{LLM-as-a-Judge metrics} reveal complementary strengths across models. \textbf{LLaMA4-Maverick} consistently achieves the highest overall scores, particularly in \textit{faithfulness}, \textit{consistency}, and \textit{citation quality}, demonstrating its balanced capability across factuality, coherence, and precision. \textbf{Gemma} performs competitively, especially on \textit{content coverage} and \textit{citation correctness}, reflecting its strong retrieval and referencing behavior. In contrast, \textbf{Mistral}, \textbf{GPT-4o}, and \textbf{DeepSeek} trail behind on most evaluation dimensions. %Overall, the narrow gaps across models suggest that while the LLM-as-a-Judge framework captures broad comparative trends, it remains limited in discerning subtle qualitative differences among high-performing systems.

\begin{table*}[htbp]
\centering
\scriptsize
\renewcommand{\arraystretch}{0.85}
\resizebox{\textwidth}{!}{
\begin{tabular}{@{}llcccccccc@{}}
\toprule
\multirow{2}{*}{\textbf{Category}} & \multirow{2}{*}{\textbf{Metric}}
& \multicolumn{4}{c}{\textbf{Instruction-Level}} & \multicolumn{4}{c}{\textbf{Few-Shot}} \\
\cmidrule(lr){3-6} \cmidrule(lr){7-10}
& & \textbf{Short} & \textbf{Medium} & \textbf{Elaborate} & \textbf{AutoCoT}
& \textbf{Zero-Shot} & \textbf{One-Shot} & \textbf{Two-Shot} & \textbf{Three-Shot} \\
\midrule
\multirow{6}{*}{Lexical Overlap}
& ROUGE-1 & 0.4261 & \textbf{0.4513} & 0.4402 & 0.4233 & 0.4402 & 0.4418 & 0.4440 & \textbf{0.4460} \\
& ROUGE-2 & 0.1622 & \textbf{0.1625} & 0.1509 & 0.1361 & 0.1509 & 0.1607 & 0.1645 & \textbf{0.1647} \\
& ROUGE-L & 0.1911 & \textbf{0.1940} & 0.1894 & 0.1701 & 0.1894 & 0.1942 & 0.1965 & \textbf{0.1972} \\
& BLEU    & 0.1117 & 0.1345 & 0.1198 & \textbf{0.1353} & 0.1198 & 0.1182 & 0.1193 & \textbf{0.1204} \\
& METEOR  & 0.2516 & 0.2797 & 0.2700 & \textbf{0.3132} & \textbf{0.2700} & 0.2669 & 0.2679 & 0.2695 \\
\midrule
\multirow{4}{*}{Semantic Similarity}
& BERTScore & \textbf{0.8480} & 0.8471 & 0.8448 & 0.8352 & 0.8448 & 0.8486 & \textbf{0.8496} & 0.8492 \\
& BLEURT & \textbf{0.3347} & 0.3312 & 0.3339 & 0.3201 & 0.3339 & 0.3431 & 0.3432 & \textbf{0.3452} \\
& Contextual Relevance & 0.9694 & 0.9690 & 0.9684 & \textbf{0.9698} & \textbf{0.9684} & 0.9677 & 0.9674 & 0.9681 \\
\midrule
\multirow{3}{*}{Content Coverage}
& Reference-based Coverage & \textbf{0.4875} & 0.4697 & 0.4845 & 0.4347 & 0.4845 & 0.4935 & 0.4913 & \textbf{0.4985} \\
& Reference-free Coverage & 0.4313 & 0.4231 & 0.4367 & \textbf{0.4556} & 0.4367 & 0.4622 & 0.4632 & \textbf{0.4755} \\
& \textit{LLM-as-a-Judge} & 0.7084 & 0.7101 & 0.7019 & \textbf{0.7340} & 0.7019 & 0.7047 & 0.7081 & \textbf{0.7107} \\
\midrule
\multirow{2}{*}{Faithfulness}
% & QA-based Faithfulness & \textbf{0.7840} & 0.7767 & 0.7794 & 0.7627 & 0.7794 & 0.7829 & 0.7780 & \textbf{0.7833} \\
% & Entailment-based Faithfulness & 0.0119 & 0.0505 & \textbf{0.1077} & -0.1368 & 0.1077 & 0.1292 & 0.1361 & \textbf{0.1432} \\
& Keyphrase-based Faithfulness & 0.2338 & 0.2183 & 0.2393 & \textbf{0.3610} & \textbf{0.2393} & 0.2353 & 0.2303 & 0.2366 \\
& \textit{LLM-as-a-Judge} & 0.8462 & 0.8382 & 0.8157 & \textbf{0.8603} & \textbf{0.8157} & 0.8077 & 0.8146 & 0.8094 \\
\midrule
\multirow{2}{*}{Consistency}
% & QA-based Consistency & \textbf{0.7870} & 0.7786 & 0.7860 & 0.7660 & 0.7860 & 0.7864 & 0.7827 & \textbf{0.7873} \\
% & Entailment-based Consistency & 0.0762 & 0.0802 & \textbf{0.1077} & -0.0521 & \textbf{0.1077} & 0.0955 & 0.0902 & 0.0852 \\
& Keyphrase-based Consistency  & 0.3450 & 0.3348 & \textbf{0.3455} & 0.2060 & 0.3455 & 0.3640 & \textbf{0.3688} & 0.3687 \\
& \textit{LLM-as-a-Judge} & 0.9045 & 0.8959 & 0.8954 & \textbf{0.9177} & 0.8954 & 0.8945 & 0.8954 & \textbf{0.8998} \\
\midrule
\multirow{4}{*}{Citation Correctness}
& Citation Recall & \textbf{0.6326} & 0.6193 & 0.5574 & 0.5350 & 0.5574 & 0.5930 & \textbf{0.6083} & 0.6076 \\
& Citation Precision & 0.8382 & 0.8548 & \textbf{0.9357} & 0.8942 & 0.9357 & 0.9506 & 0.9470 & \textbf{0.9515} \\
& \textit{LLM-as-a-Judge (Cit. Qual.)} & \textbf{0.8492} & 0.8388 & 0.8051 & 0.8243 & \textbf{0.8051} & 0.7901 & 0.7855 & 0.7823 \\
\midrule
\multirow{2}{*}{Narrative Quality}
& Perplexity $\downarrow$ & 21.4601 & 26.2687 & \textbf{18.4870} & 19.0518 & \textbf{18.4870} & 19.1426 & 23.0976 & 19.6214 \\
& \textit{LLM-as-a-Judge} & 0.8997 & 0.8950 & \textbf{0.9001} & 0.8976 & \textbf{0.9001} & 0.8999 & 0.8965 & 0.8987 \\
\bottomrule
\end{tabular}
}
\caption{Performance of LLaMA 4-Maverick on NAACL dataset under two prompting paradigms: \textit{Instruction-Level} (Short, Medium, Elaborate, AutoCoT) and \textit{Few-Shot} (Zero-Shot, One-Shot, Two-Shot, Three-Shot). Best scores in each paradigm are bolded. Note: Lower perplexity values indicate better performance.}
\label{tab:prompt-analysis}
\end{table*}

\subsection{Analysis on Few-shot Experiments}
In Table~\ref{tab:prompt-analysis}, we present the results for few-shot prompting experiments (Zero-Shot, One-Shot, Two-Shot, Three-Shot) on the NAACL dataset. The results reveal noticeable improvements with an increasing number of examples. Specifically, Three-Shot prompting consistently outperforms other few-shot settings in lexical overlap (ROUGE-1, ROUGE-2, ROUGE-L) and reference-free content coverage, suggesting that additional exemplars guide the model toward capturing finer lexical details and comprehensive content. However, improvements plateau after Two-Shot, indicating diminishing returns beyond a certain number of examples. Moreover, higher SummaC-ZS faithfulness scores in Three-Shot setups reflect that additional contexts help the model better adhere to the factual correctness of the input, highlighting the benefit of illustrative examples in complex generation tasks.

\subsection{Analysis with Various Prompt Strategies}
In Table~\ref{tab:prompt-analysis}, we present the results depicting the impact of prompt elaboration (Short, Medium, Elaborate, AutoCoT) on the NAACL dataset. Medium-length prompts achieve higher lexical overlap (ROUGE scores), suggesting that moderate complexity is optimal for balancing detail with model understanding. Conversely, AutoCoT prompts result in superior METEOR and Keyphrase-based faithfulness scores, implying that models benefit significantly from detailed reasoning steps when precision in conceptual alignment is required. However, AutoCoT notably increases perplexity, indicating potential trade-offs between detailed instructional clarity and narrative fluency. Thus, prompt complexity should be carefully adjusted depending on whether the task prioritizes lexical fidelity, conceptual alignment, or narrative coherence.

\section{Ablation Study}
\label{sec:ablation-study}
We conduct an ablation study on the overall front-runner model LLaMA4-Maverick to examine the effect of varying input configurations on the quality of generated introductions. Specifically, we compare three setups: (i) Title-only, (ii) Title + Abstract, and (iii) Title + Abstract + Related Work, under two prompting strategies: \textsc{Elaborate} and \textsc{Medium}. As shown in Table~\ref{tab:ablation}, we evaluate performance using a comprehensive set of lexical, semantic, and LLM-based metrics.

\begin{table*}[htbp]
\scriptsize
\centering
\renewcommand{\arraystretch}{0.85}
\begin{tabular}{@{}l l c c c c c c @{}}
\toprule
\multirow{2}{*}{\textbf{Category}} & \multirow{2}{*}{\textbf{Metric}} & \multicolumn{3}{c}{\textbf{Elaborate}} & \multicolumn{3}{c}{\textbf{Medium}} \\
\cmidrule(lr){3-5} \cmidrule(lr){6-8}
& & \textbf{Title-only} & \textbf{Title+Abstract} & \textbf{T+A+Related} & \textbf{Title-only} & \textbf{Title+Abstract} & \textbf{T+A+Related} \\
\midrule
\multirow{5}{*}{\textbf{Lexical Overlap}}
& ROUGE-1 & 0.3545 & 0.4093 & \textbf{0.4402} & 0.3356 & 0.3949 & \textbf{0.4513} \\
& ROUGE-2 & 0.0772 & 0.1202 & \textbf{0.1509} & 0.0660 & 0.1084 & \textbf{0.1625} \\
& ROUGE-L & 0.1553 & 0.1784 & \textbf{0.1894} & 0.1436 & 0.1700 & \textbf{0.1940} \\
& BLEU    & 0.0503 & 0.0729 & \textbf{0.1198} & 0.0223 & 0.0472 & \textbf{0.1345} \\
& METEOR  & 0.2150 & 0.2428 & \textbf{0.2700} & 0.1816 & 0.2185 & \textbf{0.2797} \\
\midrule
\multirow{3}{*}{\textbf{Semantic Similarity}}
& BERTScore & 0.8324 & 0.8408 & \textbf{0.8448} & 0.8235 & 0.8320 & \textbf{0.8471} \\
& BLEURT    & 0.3189 & 0.3333 & \textbf{0.3339} & 0.3303 & \textbf{0.3365} & 0.3312 \\
& Contextual Relevance & 0.9651 & 0.9657 & \textbf{0.9684} & 0.9633 & 0.9650 & \textbf{0.9690} \\
\midrule
\multirow{3}{*}{\textbf{Content Coverage}}
& Reference-based Coverage & 0.3918 & 0.4805 & \textbf{0.4845} & 0.3878 & \textbf{0.4715} & 0.4697 \\
& Reference-free Coverage & 0.4504 & \textbf{0.4771} & 0.4367 & 0.4602 & \textbf{0.4953} & 0.4231 \\
& \textit{LLM-as-a-Judge} & 0.5425 & 0.6916 & \textbf{0.7019} & 0.5783 & 0.6916 & \textbf{0.7101} \\
\midrule
\multirow{2}{*}{\textbf{Faithfulness}}
% & QA-based Faithfulness & 0.7662 & \textbf{0.7808} & 0.7794 & 0.7849 & \textbf{0.7858} & 0.7767 \\
% & Entailment-based Faitfulness & 0.0238 & \textbf{0.1430} & 0.1077 & -0.0854 & 0.0120 & \textbf{0.0505} \\
& Keyphrase-based Faithfulness & 0.3523 & \textbf{0.3788} & 0.3455 & 0.3495 & \textbf{0.3898} & 0.2183 \\
& \textit{LLM-as-a-Judge} & 0.7288 & \textbf{0.8557} & 0.8157 & 0.8110 & \textbf{0.8557} & 0.8382 \\
\midrule
\multirow{2}{*}{\textbf{Consistency}}
% & QA-based Consistency & 0.7780 & 0.7849 & \textbf{0.7860} & 0.7863 & \textbf{0.7889} & 0.7786 \\
% & Entailment-based Consistency & 0.0881 & \textbf{0.1261} & 0.1077 & 0.0223 & 0.0579 & \textbf{0.0802} \\
& Keyphrase-based Consistency & 0.1843 & 0.2210 & \textbf{0.2393} & 0.1780 & \textbf{0.3898} & 0.3348 \\
& \textit{LLM-as-a-Judge} & 0.7872 & \textbf{0.8977} & 0.8954 & 0.8430 & \textbf{0.8977} & 0.8959 \\
\midrule
\multirow{3}{*}{\textbf{Citation Correctness}}
& Recall & 0.0918 & 0.1199 & \textbf{0.5574} & 0.0001 & 0.0039 & \textbf{0.6193} \\
& Precision & 0.2399 & 0.3610 & \textbf{0.9357} & 0.0012 & 0.0175 & \textbf{0.8548} \\
& \textit{LLM-as-a-Judge (Cit. Qual.)} & 0.5174 & 0.5184 & \textbf{0.8051} & 0.3696 & 0.5184 & \textbf{0.8388} \\
\midrule
\multirow{2}{*}{\textbf{Narrative Quality}}
& Perplexity $\downarrow$ & \textbf{14.5978} & 19.9132 & 18.4870 & \textbf{16.6799} & 20.7426 & 26.2687 \\
& \textit{LLM-as-a-Judge} & 0.9000 & 0.8986 & \textbf{0.9001} & \textbf{0.9010} & 0.8986 & 0.8950 \\
\bottomrule
\end{tabular}
\caption{Ablation Study. Performance of LLaMA 4-Maverick on NAACL 2025 dataset with different input configurations: \textit{Title-only}, \textit{Title+Abstract}, and \textit{Title+Abstract+Related Work}. Note: Lower perplexity values indicate better performance.}
\label{tab:ablation}
\end{table*}

The results show a consistent trend: including more context leads to better
generation quality. The T+A+Related configuration achieves the best scores
across nearly all metrics and prompting strategies, with notable gains in
semantic similarity (e.g., BERTScore: 0.8448 $\rightarrow$ 0.8471), citation
precision (e.g., 0.3610 $\rightarrow$ 0.9357), and citation recall (e.g.,
0.1199 $\rightarrow$ 0.5574) when related work is added; the
size of this citation jump is expected, since overlap-based scoring penalizes
any divergence from the original citation set, and this ablation is intended
to quantify that effect rather than to treat the original citations as a
unique oracle. LLM-as-a-Judge citation quality also increases from 0.5184 to
0.8051, and the human evaluation in Section~\ref{sec:human-eval}
provides an oracle-free check on citation relevance and non-fabrication.
These findings underscore the importance of richer input, particularly related
work, in enabling LLMs to generate more accurate, well-grounded, and coherent
introductions.

% The results show a consistent trend: including more context leads to better generation quality. The T+A+Related configuration achieves the best scores across nearly all metrics and prompting strategies, with notable gains in semantic similarity (e.g., BERTScore: 0.8448 $\rightarrow$ 0.8471), citation precision (e.g., 0.3610 $\rightarrow$ 0.9357), citation recall (e.g., 0.1199 $\rightarrow$ 0.5574) when related work is added. LLM-as-a-Judge citation quality also increases from 0.5184 to 0.8051. These findings underscore the importance of richer input, particularly related work, in enabling LLMs to generate more accurate, well-grounded, and coherent introductions.

We additionally examine whether providing the full paper as context (excluding the Introduction) further sharpens generation in Appendix~\ref{app:fullcontext}, where a focused case study confirms gains in methodological detail and result-driven framing while reinforcing the realism–scalability trade-off that motivates our restricted-input design.

\section{Human Evaluation}
\label{sec:human-eval}
We conducted a human evaluation on 20 source examples, each with four model generations, LLaMA 4-Maverick (0-shot, 3-shot, AutoCoT) and GPT-4o (0-shot), totaling 80 outputs. The three authors performed the annotations themselves, each annotating 6--7 papers (four generations each). Due to the unavailability of domain expert annotators, no paper was annotated by multiple annotators. This single-annotation approach is common in recent benchmarking studies such as \citet{dhar2024can, van-der-lee-etal-2019-best} and appropriate here since the task is purely benchmarking, not proposing new models. Evaluating 50--100 outputs aligns with established LLM evaluation norms. The annotation form is included in Appendix~\ref{sec:human-eval-form}, and results are summarized in Table~\ref{tab:human-eval}.
\begin{table}[h]
\centering
\renewcommand{\arraystretch}{1}
\resizebox{\columnwidth}{!}{
\begin{tabular}{lcccc}
\toprule
\textbf{Metric} &
\makecell{\textbf{LLaMA4}\\\textbf{0-shot}} &
\makecell{\textbf{LLaMA4}\\\textbf{3-shot}} &
\makecell{\textbf{LLaMA4}\\\textbf{AutoCoT}} &
\makecell{\textbf{GPT-4o}\\\textbf{0-shot}} \\
\midrule
Faithfulness & 4.20 & \textbf{4.70} & 4.11 & 4.35 \\
Consistency & 3.80 & \textbf{4.50} & 3.84 & 4.10 \\
Content Coverage & 4.05 & \textbf{4.50} & 3.74 & 4.05 \\
Flow of Ideas & 4.30 & \textbf{4.60} & 4.26 & 4.55 \\
Citation Contextual Quality & 3.40 & \textbf{4.20} & 3.32 & 3.60 \\
Hallucination Resistance & 4.60 & \textbf{5.00} & 4.37 & 4.60 \\
Literature Context & 3.85 & \textbf{4.40} & 3.95 & 4.00 \\
Motivation Clarity & 4.15 & \textbf{4.50} & 4.21 & 4.40 \\
Methods Summary & 3.65 & \textbf{4.25} & 3.79 & 3.85 \\
Contributions Summary & 3.30 & \textbf{4.20} & 3.37 & 3.74 \\
\bottomrule
\end{tabular}
}
\caption{Human evaluation of different models across ten qualitative criteria.}
\label{tab:human-eval}
\end{table}

Across comparisons, we observe several clear trends. First, \textbf{GPT-4o outperforms 0-shot LLaMA-4} on most criteria—particularly consistency, citation contextual quality, and contribution summary—though LLaMA-4 shows stronger hallucination resistance. Second, \textbf{3-shot prompting leads to the strongest overall performance}, achieving the highest scores in content coverage, flow of ideas, and clarity of methods and contributions. Finally, \textbf{Elaborate 0-shot prompting surpasses AutoCoT} across all dimensions, with notable gains in citation quality and faithfulness, emphasizing the value of carefully designed instruction formats. These results highlight the importance of both model selection and prompting strategy in improving the quality and reliability of LLM-generated scientific introductions. 

In a detailed analysis of a NAACL 2025 dataset paper in Appendix~\ref{sec:generated-samples}, both LLaMA-4 Maverick and GPT-4o produced fluent and faithful introductions with strong narrative flow and accurate citation usage. However, the models consistently failed to recover paper-specific structure and deeper details—such as explicit research questions and contribution-style summaries—when this information was not present in the inputs, revealing a key limitation of LLM-based introduction generation.

We additionally compute pairwise Spearman correlations between the human ratings, LLM-as-Judge scores, and formula-based automated metrics; the full analysis is reported in Appendix~\ref{sec:metric-correlation}.

% Refer Appendix~\ref{sec:generated-samples} to see the generated samples and a detailed case study. In the case study with one paper from NAACL dataset and with four types of generation, we observe that the models do well in generating Literature and Motivation but not in generating Methods and Contributions Summary. Refer Appendix~\ref{sec:generated-samples} to see the generated samples and a detailed case study. In the case study with one paper from NAACL dataset and with four types of generation, we observe that the models do well in generating Literature and Motivation but not in generating Methods and Contributions Summary.

\section{Conclusion and Discussion}
Building on the findings presented in this paper, we address two central questions regarding the role of large language models (LLMs) in scientific writing. First, \textit{are LLMs good enough to generate the first draft of a research paper's introduction?} Our automated and human evaluations indicate that, when provided with structured prompts—including a clear set of instructions, a meaningful title and abstract, and a relevant set of related works—LLMs are capable of generating coherent, well-structured, and stylistically appropriate introductions that align with academic norms. Second, \textit{can the LLM-generated introduction be used as-is?} Not entirely. While LLMs produce high-quality drafts that serve as strong starting points, they fall short in several critical aspects, including the incorporation of fine-grained technical details from cited literature, accurate contextualization of citations, and precise articulation of the target paper's contributions. As such, the generated introductions require substantial post-editing and expert input. These findings suggest that LLMs are promising assistants in the scholarly writing pipeline, but they are not yet suitable replacements for human authorship.

\section*{Limitations}
\begin{enumerate}
\item In this paper, we have investigated and benchmarked LLM's capabilities in generating the text of an introduction of a research paper. However, most of the introductions in research papers also have scientific images or diagrams. Generating those diagrams is challenging since they are mostly non-natural images and abstract in nature. Thus, we have kept that out of scope for the current work and will address them in future. 
\item Our benchmark conditions generation on Title, Abstract, and Related Work only. This choice keeps the setup reproducible and scalable, but omits methodological and results-level signals that may inform certain aspects of an Introduction. Two natural extensions are (i) augmenting the input with the full Methods and Results sections, which we did not pursue here because it substantially increases token usage and relies on consistent cross-venue section extraction, or (ii) augmenting it with \emph{structured} signals distilled from those sections — for instance, automatically extracted methodological or results-level keyphrases.
We view both as concrete directions for future iterations of the benchmark.
% \item Benchmarking LLMs for generating introduction of research papers need appropriate datasets from research literature. However, such datasets get obsolete very soon due to the internet-scale training of the state-of-the-art LLMs, causing potential data leakage. This is a constant problem and thus, benchmarking datasets need to be refined on a regular basis. We believe that our approach opens up this practice for future research in this direction.
\end{enumerate}

\section*{Acknowledgement}
This research is supported in part by the NSF IIS award 2107518. Any opinions, findings, and conclusions expressed here are those of the authors and do not necessarily reflect the views of NSF. We thank our anonymous reviewers for their constructive feedback, which helped improve the quality of our paper.

\section*{Ethical Considerations}
This study investigates the capabilities of LLMs for generating scientific paper introductions, with the goal of understanding their potential as assistive tools for researchers. We explicitly do not advocate for autonomous use of LLMs in scholarly writing. While LLMs can produce coherent and stylistically appropriate drafts, our evaluation reveals persistent challenges such as factual inaccuracies, improper citation use, and limited domain grounding, all of which necessitate human validation.
We emphasize that any generated content should be reviewed and revised by human authors to ensure accuracy, integrity, and alignment with academic standards. This work is intended to benchmark current model capabilities and highlight where human involvement remains indispensable. We support responsible deployment of LLMs and caution against their use in critical academic workflows without expert oversight. We additionally caution that while proprietary LLMs outperform locally deployable models in our evaluation, their use typically requires transmitting research content to third-party servers, which raises privacy, confidentiality, and security concerns that practitioners should weigh carefully.
\bibliography{anthology-1, anthology-2, custom}

% \pagebreak

\appendix

\section*{Appendix}
\section{Justification for Model Choices}
\label{sec:model choices}
DeepSeek-V3-0324 (671B parameters, MoE) was chosen for its superior reasoning (81.2 on MMLU-Pro, outperforming GPT-4o’s score of 72.6) and ability to synthesize complex academic contexts from titles and abstracts. Llama-4-Maverick (17B active parameters) excels in processing structured academic inputs, leveraging its 1M-token context window and multimodal capabilities, critical for handling citation-rich related works. Mistral-Small-3.1-24b (24B parameters) offers efficient generation with strong academic text coherence (81\% MMLU), ideal for resource-constrained SciIG evaluation. Gemma-2-12b-it, instruction-tuned for formal prose, ensures rhetorical clarity in introductions, matching larger models on reasoning tasks. GPT-4o, a closed-source multimodal model, provides a high-performance baseline for context-heavy academic writing. These models were selected over alternatives like Claude or Qwen due to their specific strengths in reasoning, context handling, and academic prose generation, spanning diverse architectures (MoE, dense) and access types (open- and closed-source) for a comprehensive SciIG evaluation.

\section{Implementation Details}
\label{app:implementation}
This appendix documents the implementation choices behind the prompting strategies and evaluation metrics described in the main paper, with the goal of supporting reproducibility. Across all models and prompting strategies, generation uses nucleus sampling with temperature $T = 0.7$, top-$p = 0.95$, and a fixed random seed; these decoding settings are held constant within each experiment so that observed differences are attributable to the model or prompt rather than to sampling variation.

\subsection{AutoCoT Prompting}
\label{app:autocot}
Our AutoCoT strategy is implemented as a \emph{single-pass} prompting setup rather than an external iterative refinement loop. The LLM is instructed to internally generate intermediate reasoning steps (e.g., outlining the research territory, identifying the gap, and planning the contribution) before emitting the final Introduction in the same generation. We do not impose a programmatic stop condition, externally defined ``good-prompt'' criterion, or multi-round critique–revise cycle; any refinement occurs implicitly within the model's own chain of reasoning. This design isolates the effect of autonomous reasoning from additional engineering machinery and keeps the comparison across prompting strategies controlled. 

% A more structured, multi-round self-refinement framework with explicit convergence criteria is an interesting direction that we leave to future work.

\subsection{Contextual Relevance}
\label{app:ctxrel}
Contextual Relevance is computed using \texttt{allenai/longformer-base-4096} (Longformer) embeddings, chosen for their ability to encode the long Title + Abstract + cited-works context within a single forward pass. For each generated introduction and each of its associated context components (Title, Abstract, and cited works), we extract the last-hidden-state representations and apply mask-aware mean pooling over the non-padding tokens to obtain a fixed-dimensional vector. Contextual Relevance is then the mean of the cosine similarities between the generated-introduction embedding and the embeddings of its context components. This metric does \emph{not} rely on BERTScore or Sentence-BERT, both of which we use elsewhere only for different evaluation components.

\subsection{Keyphrase Extraction and Matching}
\label{app:kbert}
Keyphrases are extracted using KeyBERT\footnote{\url{https://github.com/MaartenGr/KeyBERT}} with the \texttt{all-MiniLM-L6-v2} Sentence-BERT backbone. For each text (generated introduction, ground-truth introduction, and the Title/Abstract input context), we extract the top-5 keyphrases with an n-gram range of (1, 3), using English stopword removal. Matching is performed via exact equality after stemming with the Porter stemmer. The same procedure is used for both Reference-based Coverage (matching against ground-truth introduction keyphrases) and Reference-free Coverage (matching against keyphrases extracted from the Title and Abstract), ensuring that the two coverage metrics differ only in the reference set, not in the matching procedure.

\begin{table*}[t]
\centering
\small
\begin{tabular}{lrr}
\toprule
\textbf{Stage / Outcome} & \textbf{Count} & \textbf{\%} \\
\midrule
\multicolumn{3}{l}{\textit{Citation-level lookup resolution}} \\
\midrule
Total related-paper lookups            & 59{,}619 & 100.0 \\
\quad Stage 1: unofficial endpoint     & 55{,}207 &  92.6 \\
\quad Stage 2: official Graph API fallback &    775 &   1.3 \\
\textbf{Total resolved}                & \textbf{55{,}982} & \textbf{94.0}\rlap{$^{\dagger}$} \\
Unmatched after both stages            &  3{,}279 &   5.5 \\
\midrule
\multicolumn{3}{l}{\textit{Metadata completeness among resolved}} \\
\midrule
Complete (title + abstract + authors)  & 50{,}384 &  90.0 \\
Missing abstract (title + authors only)&  5{,}598 &  10.0 \\
\midrule
\multicolumn{3}{l}{\textit{Target-paper outcome}} \\
\midrule
Target papers excluded (all citations unresolved) & 150 & --- \\
\bottomrule
\end{tabular}
\caption{Semantic Scholar metadata resolution outcomes across the two datasets. Stage~1 and Stage~2 percentages are reported with respect to the total number of related-paper lookups; metadata-completeness percentages are reported with respect to the resolved subset. $^{\dagger}$Stage~1 and Stage~2 sum to 93.9\% due to rounding.}
\label{tab:s2resolution}
\end{table*}

\section{Semantic Scholar Metadata Resolution}
\label{app:s2resolution}
The Semantic Scholar API occasionally returns \texttt{None} for individual fields (most commonly the abstract or author list) when a cited paper is poorly indexed or newly published. To handle these cases consistently and avoid silently dropping records, our Step~5 metadata retrieval follows a three-stage strategy:

\begin{enumerate}
    \item \textbf{Unofficial endpoint with title normalization.} Each cited title is first queried through the unofficial Semantic Scholar endpoint, using both the original title string and a normalized variant (lowercased, punctuation- and whitespace-collapsed).
    \item \textbf{Official Graph API fallback with similarity matching.} If abstract or author fields remain missing after Stage~1, we fall back to the official Semantic Scholar Graph API \citep{Kinney2023TheSS}, accepting a returned record only if its title matches the queried title with a similarity score $\geq 0.85$, in order to avoid spurious associations to near-duplicate titles.
    \item \textbf{Exclusion of fully unresolved citations.} Related papers for which neither endpoint returns usable metadata are excluded from the citation context. If \emph{all} cited papers for a target paper remain unresolved, that target paper is removed from the dataset entirely.
\end{enumerate}

Across the two datasets, we processed 59,619 related-paper lookups. Overall, 94.0\% of citations were successfully resolved through the two-stage retrieval process, with the vast majority obtained from the unofficial endpoint and a smaller fraction recovered through the official Graph API fallback. Among resolved entries, roughly 10\% lacked abstracts in Semantic Scholar's index but still contributed title and author metadata. Target papers for which all cited references remained unresolved were excluded from the final dataset. Table~\ref{tab:s2resolution} summarizes the full resolution statistics.

\section{Metric Agreement Analysis}
\label{sec:metric-correlation}
To assess how the three evaluation channels relate, we compute Spearman
rank correlations between human ratings, the LLM-as-Judge scores, and
the corresponding \emph{formula-based} automated metrics (e.g.,
keyphrase-based Faithfulness, reference-based Coverage, citation Recall,
$-$Perplexity) on the human-evaluated subset ($n{=}79$--$80$;
Table~\ref{tab:correlation-summary}). LLM-as-Judge agrees
\emph{moderately} with the formula-based metrics across Faithfulness
($\rho{=}0.35$), Consistency ($\rho{=}0.45$), Content Coverage
($\rho{=}0.30$), and Narrative Quality ($\rho{=}0.34$), supporting its
use as an automatic surrogate for these dimensions. The one exception is
citation quality ($\rho{=}-0.19$), where the LLM-judge and the
overlap-based citation metric diverge --- consistent with our earlier
observation (Section~\ref{sec:ablation-study}) that overlap penalizes
contextually valid but non-identical citations. Both the formula-based
metrics and LLM-as-Judge show only weak agreement with human ratings
($|\rho| \le 0.20$), in line with known limitations of automatic
generation metrics; this motivates our multi-channel evaluation design
rather than reliance on any single metric family.

\begin{table}[h]
\centering
\small
\resizebox{\columnwidth}{!}{%
\begin{tabular}{lccc}
\toprule
\textbf{Dimension} & \textbf{H--Formula} & \textbf{LLM-J--Formula} & \textbf{H--LLM-J} \\
\midrule
Faithfulness       & $+0.08$ & $+0.35$ & $+0.03$ \\
Consistency        & $-0.10$ & $+0.45$ & $+0.05$ \\
Content Coverage   & $+0.12$ & $+0.30$ & $-0.15$ \\
Citation           & $+0.06$ & $-0.19$ & $+0.20$ \\
Fluency/Narrative  & $-0.05$ & $+0.34$ & $+0.04$ \\
\bottomrule
\end{tabular}
}
\caption{Spearman rank correlations across the three evaluation channels
on the human-evaluated subset ($n{=}79$--$80$). \textbf{H}: Human;
\textbf{LLM-J}: LLM-as-Judge; \textbf{Formula}: formula-based automated metric.
Interpretive bands: $|\rho| < 0.10$ negligible, $0.10$--$0.30$ weak,
$0.30$--$0.50$ moderate. Human scores are on a $1$--$5$ ordinal scale; LLM-Judge scores
on a $0$--$1$ continuous scale.}
\label{tab:correlation-summary}
\end{table}

\section{Dataset Construction Pipeline: Detailed Steps}
\label{app:pipeline}

This appendix expands the five-step dataset construction pipeline summarized in Section~\ref{sec:datasets} and illustrated in Figure~\ref{fig:pipeline}. The full procedure for handling partial Semantic Scholar API responses is described separately in Appendix~\ref{app:s2resolution}.

\paragraph{Step 1: Retrieval.}
We retrieve PDF files of accepted NAACL 2025 and ICLR 2025 papers from web sources, including official conference proceedings and repositories like ArXiv. For each conference, we aggregated all available PDFs to ensure comprehensive coverage of that venue's publications.

\paragraph{Step 2: Parsing.}
We process the PDFs using the \texttt{grobid2json} parsing tool provided by the S2ORC project \citep{lo-etal-2020-s2orc}. This tool internally leverages the Grobid library to first convert PDFs into structured TEI-XML. Subsequently, it extracts essential components such as titles, abstracts, introductions, author details, and bibliographic references, and organizes them into a structured JSON format. This structured representation provides a robust foundation for the subsequent steps.

\paragraph{Step 3: Citation Extraction.}
We develop a regular expression to identify Related Works cited within each paper's introduction. The regex targets citation patterns (e.g., ``Author et al., Year'') to extract references relevant to the introductory context, capturing the scholarly connections embedded in the text.

\paragraph{Step 4: Citation Mapping.}
We employ the \textbf{LLaMA 4-Maverick} model to map the extracted Related Works citations (e.g., ``et al.'' strings) to entries in the references section, retrieving the corresponding titles. The prompt used for this citation mapping is provided in Table~\ref{tab:citation-mapping-prompt}. This step ensures accurate linkage between in-text citations and their full bibliographic details.

\paragraph{Step 5: Enrichment.}
We use the SemanticScholar API, combining the official\footnote{\url{https://www.semanticscholar.org/}} \cite{Kinney2023TheSS} and unofficial\footnote{\url{https://github.com/danielnsilva/semanticscholar}} endpoints, to fetch abstracts and author details for each Related Works title. This enriches the dataset with contextual citation information, supporting tasks requiring citation-aware generation. The multi-stage resolution strategy and handling of partial \texttt{None} responses are detailed in Appendix~\ref{app:s2resolution}.

\section{Evaluation Metric Details}
\label{app:eval-details}

This appendix expands the per-metric definitions summarized in Section~\ref{sec:eval}. Implementation specifics for Contextual Relevance and keyphrase extraction are documented separately in Appendices~\ref{app:ctxrel} and~\ref{app:kbert}; the keyphrase-based Faithfulness formulation is detailed in Appendix~\ref{sec:appendix-faithfulness}.

\subsection{Lexical Overlap}
We use five standard surface-level metrics:
\begin{itemize}
    \item \textbf{ROUGE-1} and \textbf{ROUGE-2} compute the unigram and bigram overlap respectively between the generated and ground-truth introductions, expressed as the ratio of matching $n$-grams (balancing precision and recall).
    \item \textbf{ROUGE-L} measures the longest common subsequence, capturing structural similarity at sentence and discourse granularity.
    \item \textbf{BLEU} computes modified $n$-gram precision with a brevity penalty for length mismatch between candidate and reference.
    \item \textbf{METEOR} aligns words by stemming and synonymy, making the score sensitive to morphological and lexical variation that ROUGE and BLEU would miss.
\end{itemize}

\subsection{Semantic Similarity}
\textbf{BERTScore} computes cosine similarity between contextual token embeddings of the generated and ground-truth introductions using a pretrained BERT model. Because the embeddings are contextual, paraphrastic and syntactic variation are matched softly, unlike $n$-gram overlap.

\textbf{BLEURT} uses a BERT-based regression model fine-tuned on human ratings of generation quality, producing a single similarity score per (candidate, reference) pair. It is generally more sensitive to fluency and adequacy than BERTScore.

\textbf{Contextual Relevance} is defined formally in Section~\ref{sec:eval}: the mean cosine similarity of the generated-introduction embedding with the embeddings of its input context (title, abstract, and each cited paper). Because the score is averaged over a heterogeneous set of context elements, it is robust to either (a) one element drifting in topic or (b) one element being very short. The embedding model and pooling strategy are detailed in Appendix~\ref{app:ctxrel}.

% \subsection{Content Coverage}
% Reference-based Coverage (RBC) is defined in Section~\ref{sec:eval}. Reference-free Coverage (RFC) is defined analogously over the input-context keyphrase set $\mathcal{K}_{\text{ctx}}$:
% \[
% \text{RFC} = \frac{|\{k \in \mathcal{K}_{\text{ctx}} \mid \exists \, k' \in \mathcal{K}_{\text{gen}},\, k \sim k'\}|}{|\mathcal{K}_{\text{ctx}}|}.
% \]
% RBC and RFC share the same keyphrase extractor and the same three-stage semantic-match procedure (Porter-stemmed exact match, partial token overlap, Sentence-BERT cosine similarity at $\tau{=}0.7$); they differ only in the reference set. The two are complementary: RBC measures recovery of the published introduction's key content, while RFC measures grounding in the explicit input context.

\section{Faithfulness and Consistency Metrics}
\label{sec:appendix-faithfulness}

\textbf{Faithfulness} assesses the factual alignment between the generated Introduction and its source inputs (Title and Abstract), and \textbf{Consistency} assesses alignment with the published ground-truth introduction. For each dimension we report one keyphrase-based automatic metric and one LLM-as-a-Judge score (GPT-4o; prompts in Table~\ref{tab:llm-judge-prompts}). Both keyphrase-based metrics share the extraction and semantic-matching procedure used by RBC and RFC (Appendix~\ref{app:kbert}).

\textsc{Keyphrase-based Faithfulness.} Let $\mathcal{K}_{\text{gen}}$ be the set of keyphrases extracted from the generated introduction, and $\mathcal{K}_{\text{ctx}}$ the set extracted from the Title and Abstract (serving as factual ground truth). The two keyphrases satisfy the exact match $k = k'$ only if their stemmed forms are identical.
\[
\text{Faithfulness}_{\text{KP}} = \frac{|\{k \in \mathcal{K}_{\text{gen}} \mid \exists \, k' \in \mathcal{K}_{\text{ctx}}, \, k = k'\}|}{|\mathcal{K}_{\text{gen}}|}.
\]
This captures the proportion of generated keyphrases that are factually grounded in the source context.

\textsc{Keyphrase-based Consistency.} Consistency uses the same formulation but with the ground-truth introduction's keyphrase set $\mathcal{K}_{\text{GT}}$ as the reference instead of $\mathcal{K}_{\text{ctx}}$:
\[
\text{Consistency}_{\text{KP}} = \frac{|\{k \in \mathcal{K}_{\text{gen}} \mid \exists \, k' \in \mathcal{K}_{\text{GT}}, \, k = k'\}|}{|\mathcal{K}_{\text{gen}}|}.
\]
It assesses whether the generated text maintains logical and factual consistency with the published introduction.

\subsection{Citation Correctness}
Citation Precision and Citation Recall operate over citation \emph{identifiers} (``et al.'' strings) rather than over titles or DOIs; identifiers are extracted from the generated introduction using LLaMA 4-Maverick with the prompt in Table~\ref{tab:citation-mapping-prompt}. Precision penalizes citations that do not appear in the paper's actual Related Works list (potentially fabricated or off-topic), while Recall penalizes failure to incorporate citations that the paper's introduction did use. \textbf{Citation Contextual Quality} (LLM-as-a-Judge) is independent of overlap with the original citation set; it scores whether any cited sentence accurately reflects the cited paper's metadata (title, abstract, authors) and is placed in a contextually appropriate position.

\subsection{Narrative Quality}
Perplexity is computed under GPT-2 \citep{radford2019language} as
\[
\text{PPL} = \exp\!\Big(\tfrac{1}{N} \sum_{i=1}^{N} -\log p(w_i \mid w_{<i})\Big),
\]
where $w_1, \dots, w_N$ is the tokenized generated introduction. Lower values indicate generations that the language model finds more probable, used here as a proxy for surface fluency. The LLM-as-a-Judge Narrative Quality score (GPT-4o) complements perplexity by grading higher-order properties — grammar, logical flow between paragraphs, academic tone, and overall readability — that token-level predictability does not capture.

\subsection{LLM-as-a-Judge}
All LLM-as-a-Judge scores use GPT-4o with structured JSON-output prompts (Table~\ref{tab:llm-judge-prompts}). Each prompt elicits both a numeric score in $[0, 1]$ and a short bullet-style rationale; we use the numeric score in Table~\ref{tab:results-main} and reserve the rationale for qualitative inspection and case studies. Using a single judge model across all five LLM-judge dimensions (Content Coverage, Faithfulness, Consistency, Citation Contextual Quality, Narrative Quality) keeps the scoring rubric internally consistent.

\begin{table*}[htbp]
\centering
\footnotesize
\renewcommand{\arraystretch}{1.05}
\begin{tabular}{@{}p{3.8cm}p{12.5cm}@{}}
\toprule
\textbf{Metric} & \textbf{LLM Prompt (User Message)} \\
\midrule
\textbf{Content Coverage \& Consistency} &
\begin{minipage}[t]{\linewidth}
\texttt{Generated Introduction:} \newline
\texttt{""" \{pred\} """} \newline
\texttt{Actual Introduction:} \newline
\texttt{""" \{label\} """} \newline
\texttt{You will evaluate a Generated Introduction against the Actual Introduction and output ONLY this JSON:} \newline
\texttt{\{ "Content Coverage": "float (0.0–1.0)",} \newline
\texttt{  "Content Coverage Rationale": "bullet-style justification",} \newline
\texttt{  "Consistency": float (0.0–1.0),} \newline
\texttt{  "Consistency Rationale": "bullet-style justification" \}} \newline
\texttt{---} \newline
\texttt{Scale: 0.0 = no alignment, 1.0 = perfect match} \newline
Content Coverage checks how many core claims and supporting facts from the Actual Introduction appear in the Generated Introduction. \newline
Consistency checks for hallucinations or internal contradictions. Partial scores are allowed.
\end{minipage} \\
\midrule
\textbf{Faithfulness \& Citation Context Quality} &
\begin{minipage}[t]{\linewidth}
\texttt{Generated Introduction:} \newline
\texttt{""" \{pred\} """} \newline
\texttt{Metadata: \{Title, Abstract, Related Papers\}} \newline
\texttt{You are evaluating a Generated Introduction against its Metadata and output ONLY this JSON:} \newline
\texttt{\{ "Faithfulness": "float (0.0–1.0)",} \newline
\texttt{  "Faithfulness Rationale": "bullet-style justification",} \newline
\texttt{  "Citation Context Quality": "float (0.0–1.0)",} \newline
\texttt{  "Citation Context Quality Rationale": "bullet-style justification" \}} \newline
Faithfulness checks if each claim is supported by the Title, Abstract, or cited papers. Citation Quality checks if cited sentences accurately reflect cited metadata.
\end{minipage} \\
\midrule
\textbf{Narrative Quality} &
\begin{minipage}[t]{\linewidth}
\texttt{Generated Introduction:} \newline
\texttt{""" \{pred\} """} \newline
\texttt{Scale: 0.0 = disfluent, 1.0 = highly fluent and readable} \newline
\texttt{Output ONLY JSON:} \newline
\texttt{\{ "Narrative Quality": "float (0.0–1.0)",} \newline
\texttt{  "Narrative Quality Rationale": "bullet-style justification" \}} \newline
Evaluation includes grammar, logical flow, academic tone, and readability. Deduct points for issues; provide justification in bullet style.
\end{minipage} \\
\bottomrule
\end{tabular}
\caption{Prompts used for LLM-as-a-Judge evaluation. Prompts are structured to elicit scores and justifications across five quality dimensions.}
\label{tab:llm-judge-prompts}
\end{table*}
\begin{table*}[htbp]
\centering
\renewcommand{\arraystretch}{1.4}
\resizebox{\textwidth}{!}{
\begin{tabular}{@{}clp{7.5cm}c p{5.5cm}@{}}
\toprule
\textbf{\#} & \textbf{Criterion} & \textbf{Guiding Questions} & \textbf{Rating (1–5 Stars)} & \textbf{Comments} \\
\midrule
\multicolumn{5}{@{}l}{\textbf{Content Quality}} \\
\midrule
1 & Faithfulness & Does the generated Introduction accurately reflect information from the Title, Abstract, and Related Papers without introducing facts not supported by them? & $\star$ $\star$ $\star$ $\star$ $\star$ & \\
2 & Consistency & Is the narrative consistent with the Ground Truth Introduction in terms of logical structure, intent, and internal coherence? Are there any contradictions in meaning or claims? & $\star$ $\star$ $\star$ $\star$ $\star$ & \\
3 & Coverage & Does the Introduction include all key concepts and major points found in the Title, Abstract, and Related Papers? Are critical topics missing or underrepresented? & $\star$ $\star$ $\star$ $\star$ $\star$ & \\
4 & Flow of Ideas & Does the text follow a clear and logical progression from background to motivation, method, and contributions? Are transitions smooth and well-structured? & $\star$ $\star$ $\star$ $\star$ $\star$ & \\
5 & Citation Contextual Quality & Are citations used appropriately to support specific points (e.g., prior work, problem framing, methodology)? Are they integrated fluently into the narrative? & $\star$ $\star$ $\star$ $\star$ $\star$ & \\
6 & Hallucinated Citations & Are any citations fabricated, irrelevant, or inaccurately attributed? & $\star$ $\star$ $\star$ $\star$ $\star$ & \\
\midrule
\multicolumn{5}{@{}l}{\textbf{Narrative Quality}} \\
\midrule
7 & Background / Literature Context & Is the background informative and relevant? Are related works used to effectively contextualize the research? & $\star$ $\star$ $\star$ $\star$ $\star$ & \\
8 & Motivation / Problem Statement & Is the research motivation clearly presented? Is the need or gap in existing literature convincingly established? & $\star$ $\star$ $\star$ $\star$ $\star$ & \\
9 & Method / Approach Summary & Is the proposed method succinctly and accurately described? Does it clearly build upon or differ from prior work? & $\star$ $\star$ $\star$ $\star$ $\star$ & \\
10 & Contributions Summary & Are the contributions of the paper stated clearly and aligned with the core research claims? Are they appropriately emphasized? & $\star$ $\star$ $\star$ $\star$ $\star$ & \\
\bottomrule
\end{tabular}}
\caption{Human evaluation rubric for assessing the quality of generated Introductions. Metrics are grouped under content and narrative quality. Each row includes a guiding question, a rating scale, and space for evaluator comments.}
\label{sec:human-eval-form}
\end{table*}
\begin{table*}[htbp]
\centering
\scriptsize
\renewcommand{\arraystretch}{1.15}
\begin{tabular}{@{}p{13cm}@{}}
\toprule
\textbf{LLaMA 4-Maverick Citation Mapping Prompt} \\
\midrule
\textbf{Objective:} Extract citations from paper introductions using \textbf{LLaMA 4-Maverick}, returning a JSON object mapping citation identifiers to titles and reasons for inclusion. Respond only with the JSON object, without extra text. \\
\midrule
\textbf{Paper Metadata:} \\

Title: \{title\} \\
Abstract: \{abstract\} \\
Authors: \{authors\} \\
\midrule
\textbf{Introduction Data:} \\
Full introduction text. \\
Provided citation records: \\
\quad - Citation \{paper\_id\}: \\
\quad\quad - Title: \{title\} \\
\quad\quad - Authors: \{authors\} \\
\quad\quad - Full Reference: \{raw\_text\} \\
\midrule
\textbf{Task:} \\

Prioritize the high-confidence citation list, then scan the introduction for additional citations. \\
For each citation, create an object with: \\
\quad - \texttt{title}: The citation's title. \\
\quad - \texttt{reason}: \\
\quad\quad - If in the high-confidence list: ``This citation is present in high-confidence list.'' \\
\quad\quad - Otherwise: ``Not included in the high-confidence list but cited in Introduction as (\{Author et al., Year\}).'' \\
Return a JSON object with a single key \texttt{neighbors}, containing an array of citation objects. \\
Each citation object maps the identifier (e.g., ``Li et al., 2022'') to a \texttt{title} and \texttt{reason} pair. \\
Do not emit bare arrays, Markdown, comments, or additional text—only the JSON object. \\
\midrule
\textbf{Required Output Example:} \\
\begin{verbatim}
{
"neighbors": [
{
"Li et al., 2022": {
"title": "Convergence of Adam under relaxed assumptions",
"reason": "This citation is present in high-confidence list"
}
},
{
"Hashimoto et al., 2018": {
"title": "Fairness without demographics in repeated loss minimization",
"reason": "This citation is present in high-confidence list"
}
},
{
"Blitzer et al., 2006": {
"title": "Domain adaptation with structural correspondence learning",
"reason": "Not included in the high-confidence list but cited in Introduction as (Blitzer et al., 2006)"
}
}
]
}
\end{verbatim} \\
\bottomrule
\end{tabular}
\caption{Prompt for citation extraction (Step 4) using LLaMA 4-Maverick, avoiding invalid outputs like bare arrays.}
\label{tab:citation-mapping-prompt}
\end{table*}

\begin{table*}[htbp]
\small
\centering
\renewcommand{\arraystretch}{1.05}
\resizebox{\textwidth}{!}{
\begin{tabular}{@{}lp{16cm}@{}}
\toprule
\textbf{Prompt Level} & \textbf{Prompt} \\
\midrule
\textbf{Short} & You are an AI assistant tasked with generating the Introduction section of a research paper. Below, you are provided with the Title and Abstract of the TARGET PAPER, along with details of RELATED PAPERS that must be cited within the Introduction. Please craft an academic-style Introduction incorporating these citations appropriately.

TARGET PAPER:\newline
Title: \{title\}\newline
Abstract: \{abstract\}

RELATED PAPERS:\newline
\{cited\_nodes\_pretty\}

Introduction: \\
\midrule
\textbf{Medium} & You are an AI assistant tasked with generating a structured and well-contextualized Introduction section of a research paper. Below, you will find the Title and Abstract of the TARGET PAPER. Additionally, details of several RELATED PAPERS are provided, which must be cited appropriately within the Introduction. Please incorporate these citations naturally, highlighting how the target paper builds upon or differentiates itself from previous works.

IMPORTANT FORMAT REQUIREMENTS:\newline

Clearly establish the research context and significance.\newline
Identify and articulate the research gap or question addressed by the target paper.\newline
Highlight how the contributions of the target paper relate to or extend the cited related works.\newline
Maintain formal academic writing standards.

TARGET PAPER:\newline
Title: \{title\}\newline
Abstract: \{abstract\}

RELATED PAPERS:\newline
Below is a JSON dictionary listing related papers. Each key represents a citation identifier (e.g., 'Li et al. 2022'). Each entry includes the title, authors, and abstract of the paper. The papers are provided in no specific order, and you should incorporate citations naturally where contextually relevant.\newline
\{cited\_nodes\_pretty\}

Introduction: \\
\midrule
\textbf{Elaborate} & You are an AI assistant tasked with generating a detailed and structured Introduction section of a research paper based on the provided Title and Abstract. Below, you will find comprehensive descriptions of the TARGET PAPER and several RELATED PAPERS. The TARGET PAPER's Abstract outlines its primary objectives, methods, and potential contributions. The RELATED PAPERS provide crucial background information, research developments, and existing gaps in the field. Integrate insights from these RELATED PAPERS effectively to establish a clear research context, articulate the significance of existing gaps, and explicitly highlight how the TARGET PAPER addresses these gaps through its novel contributions. Cite these papers in-text using APA format wherever applicable.

IMPORTANT FORMAT REQUIREMENTS:\newline

Your response MUST consist of EXACTLY FOUR PARAGRAPHS for the Introduction.\newline
DO NOT deviate from this four-paragraph structure.\newline
Each paragraph must be between 100-150 words, totaling approximately 600 words.

STRUCTURE:\newline

Paragraph 1: Broad overview of the research area, contextual insights from RELATED PAPERS, significance of the topic.\newline
Paragraph 2: Specific problem or gap identified, supported by RELATED PAPERS.\newline
Paragraph 3: Novel contributions of the TARGET PAPER, comparison to RELATED PAPERS.\newline
Paragraph 4: Summary of significance, potential impact, and research purpose.

STYLE AND CONTENT REQUIREMENTS:\newline

Maintain formal academic tone.\newline
Coherent and concise writing directly related to the Title and Abstract.\newline
Effective use of transitional phrases.

CITATION INSTRUCTIONS:\newline

DO NOT invent or hallucinate citations.\newline
ONLY cite the provided RELATED PAPERS.\newline
Use APA in-text citation format, e.g., '(Smith et al.)'.

TARGET PAPER:\newline
Title: \{title\}\newline
Abstract: \{abstract\}

RELATED PAPERS:\newline
Below is a JSON dictionary listing related papers. Each key represents a citation identifier (e.g., 'Li et al. 2022'). Each entry includes the title, authors, and abstract of the paper. The papers are provided in no specific order, and you should incorporate citations naturally where contextually relevant.\newline
\{cited\_nodes\_pretty\}

Introduction: \\
\bottomrule
\end{tabular}}
\caption{Prompting strategies (Part 1): Short, Medium, and Elaborate.}
\label{tab:prompts-1}
\end{table*}
\begin{table*}[htbp]
\small
\centering
\renewcommand{\arraystretch}{1.05}
\resizebox{\textwidth}{!}{
\begin{tabular}{@{}lp{16cm}@{}}
\toprule
\textbf{Prompt Level} & \textbf{Prompt} \\
\midrule
\textbf{AutoCoT} & You are an advanced AI assistant performing AutoCoT prompting to autonomously generate intermediate prompts leading to the final Introduction section of a research paper. You have the Title and Abstract of the TARGET PAPER, detailing its scope and contributions, and comprehensive information about RELATED PAPERS, providing necessary background and context.

GUIDELINES:\newline

First, autonomously generate and iteratively refine intermediate prompts to:\newline
a. Summarize research context and significance from RELATED PAPERS.\newline
b. Clearly identify gaps or unresolved challenges.\newline
c. Outline the unique contributions and relevance of the TARGET PAPER.\newline
Next, use these refined intermediate prompts to craft a coherent, structured Introduction.

STRUCTURE REQUIREMENTS:\newline

The Introduction must contain exactly four paragraphs, each 100–150 words:\newline

Contextual overview and significance of the topic.\newline
Clearly defined research gap, supported by RELATED PAPERS.\newline
Novel contributions of the TARGET PAPER, contrasting with existing literature.\newline
Summary of significance, potential impacts, and thesis statement.

FORMAT AND CITATION REQUIREMENTS:\newline

Maintain academic rigor, formal tone, and smooth transitions.\newline
Accurately cite only the provided RELATED PAPERS using APA in-text citation format (e.g., '(Smith et al.)').

TARGET PAPER:\newline
Title: \{title\}\newline
Abstract: \{abstract\}

RELATED PAPERS:\newline
Below is a JSON dictionary listing related papers. Each key represents a citation identifier (e.g., 'Li et al. 2022'). Each entry includes the title, authors, and abstract of the paper. The papers are provided in no specific order, and you should incorporate citations naturally where contextually relevant.\newline
\{cited\_nodes\_pretty\}

Introduction: \\
\bottomrule
\end{tabular}}
\caption{Prompting strategies (Part 2): AutoCoT.}
\label{tab:prompts-2}
\end{table*}
\begin{table*}[htbp]
\small
\centering
\renewcommand{\arraystretch}{1.1}
\resizebox{\textwidth}{!}{
\begin{tabular}{@{}lp{16cm}@{}}
\toprule
\textbf{Prompt Level} & \textbf{Prompt} \\
\midrule
\textbf{ZERO\_SHOT} & Same as the Elaborate prompt in Table~\ref{tab:prompts-1} \\
\hline
\textbf{One-Shot} & You are an AI assistant tasked with generating a detailed and structured Introduction section of a research paper based on the provided Title and Abstract. Below, you will find comprehensive descriptions of the TARGET PAPER and several RELATED PAPERS. The TARGET PAPER's Abstract outlines its primary objectives, methods, and potential contributions. The RELATED PAPERS provide crucial background information, research developments, and existing gaps in the field. Integrate insights from these RELATED PAPERS effectively to establish a clear research context, articulate the significance of existing gaps, and explicitly highlight how the TARGET PAPER addresses these gaps through its novel contributions. Cite these papers in-text using APA format wherever applicable.

IMPORTANT FORMAT REQUIREMENTS:\newline

Your response MUST consist of EXACTLY FOUR PARAGRAPHS for the Introduction.\newline
DO NOT deviate from this four-paragraph structure.\newline
Each paragraph must be between 100-150 words, totaling approximately 600 words.

STRUCTURE:\newline

Paragraph 1: Broad overview of the research area, contextual insights from RELATED PAPERS, significance of the topic.\newline
Paragraph 2: Specific problem or gap identified, supported by RELATED PAPERS.\newline
Paragraph 3: Novel contributions of the TARGET PAPER, comparison to RELATED PAPERS.\newline
Paragraph 4: Summary of significance, potential impact, and research purpose.

STYLE AND CONTENT REQUIREMENTS:\newline

Maintain formal academic tone.\newline
Coherent and concise writing directly related to the Title and Abstract.\newline
Effective use of transitional phrases.

CITATION INSTRUCTIONS:\newline

DO NOT invent or hallucinate citations.\newline
ONLY cite the provided RELATED PAPERS.\newline
Use APA in-text citation format, e.g., '(Smith et al.)'.

Here is an example showing how to generate an Introduction (output) based on a paper's Title, Abstract, and Related Papers (inputs).

[Insert one example here.]

Now, based on the above example and the guidelines below, please generate the Introduction for the following TARGET PAPER.

TARGET PAPER:\newline
Title: \{title\}\newline
Abstract: \{abstract\}\newline
RELATED PAPERS:\newline
\{cited\_nodes\_pretty\}\newline
Introduction: \\
\hline
\textbf{Two-Shot} & Here are two examples showing how to generate an Introduction (output) based on a paper's Title, Abstract, and Related Papers (inputs).

[Insert two examples here.]

Now, based on the above examples and the guidelines below, please generate the Introduction for the following TARGET PAPER.

[Elaborate prompt continues as above.] \\
\hline
\textbf{Three-Shot} & Here are three examples showing how to generate an Introduction (output) based on a paper's Title, Abstract, and Related Papers (inputs).

[Insert three examples here.]

Now, based on the above examples and the guidelines below, please generate the Introduction for the following TARGET PAPER.

[Elaborate prompt continues as above.] \\
\bottomrule
\end{tabular}}
\caption{Prompting strategies (Part 3): Few-shot prompting with 0–3 examples prepended to the Elaborate prompt.}
\label{tab:prompts-3}
\end{table*}

\section{A Case Study}
\label{sec:generated-samples}
\begin{figure*}[t]
  \centering
  \includegraphics[width=\linewidth]{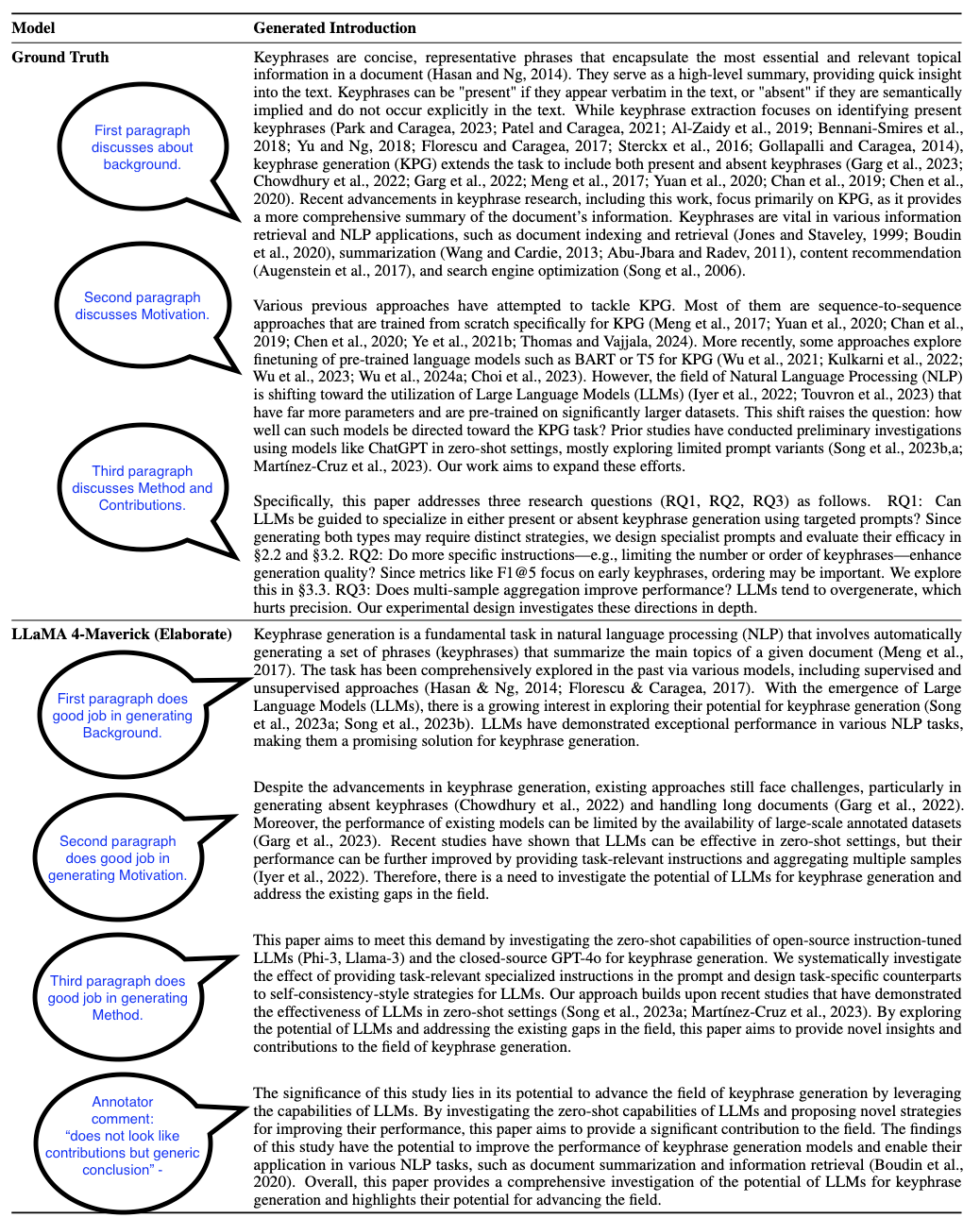}
  \caption{Ground Truth and Generated Introduction by LLaMA 4-Maverick (Elaborate) for \textit{Zero-Shot Keyphrase Generation: Investigating Specialized Instructions and Multi-Sample Aggregation on Large Language Models}.}
  \label{tab:gen-gt}
\end{figure*}

\begin{figure*}[t]
  \centering
  \includegraphics[width=\linewidth]{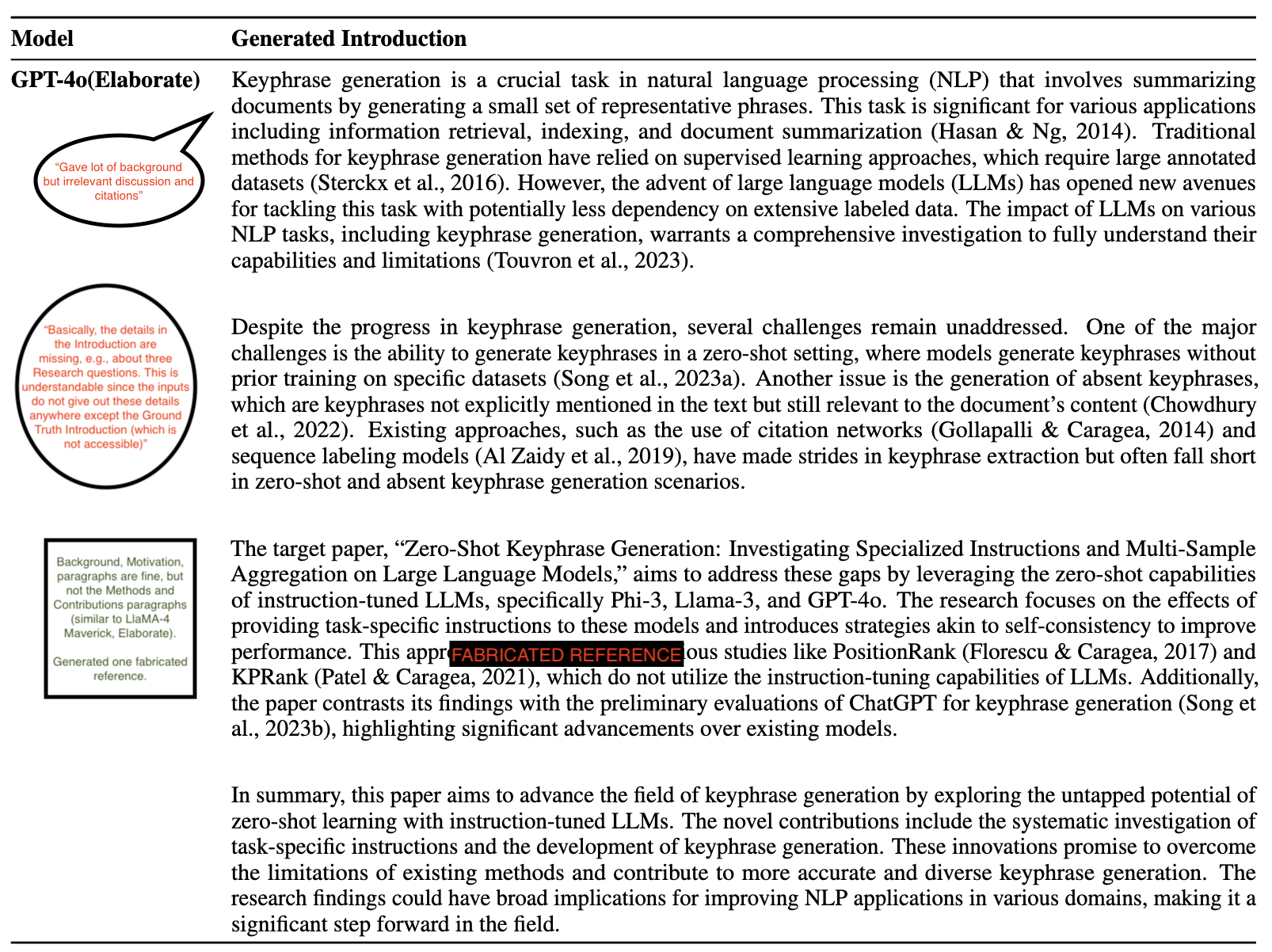}
  \caption{Ground Truth and Generated Introduction by LLaMA 4-Maverick (Elaborate) for \textit{Zero-Shot Keyphrase Generation: Investigating Specialized Instructions and Multi-Sample Aggregation on Large Language Models}. The comment in red is Annotators' comment, whereas the comment in green is our comment.}
  \label{tab:gen-gpt4o}
\end{figure*}

\begin{figure*}[t]
  \centering
  \includegraphics[width=\linewidth]{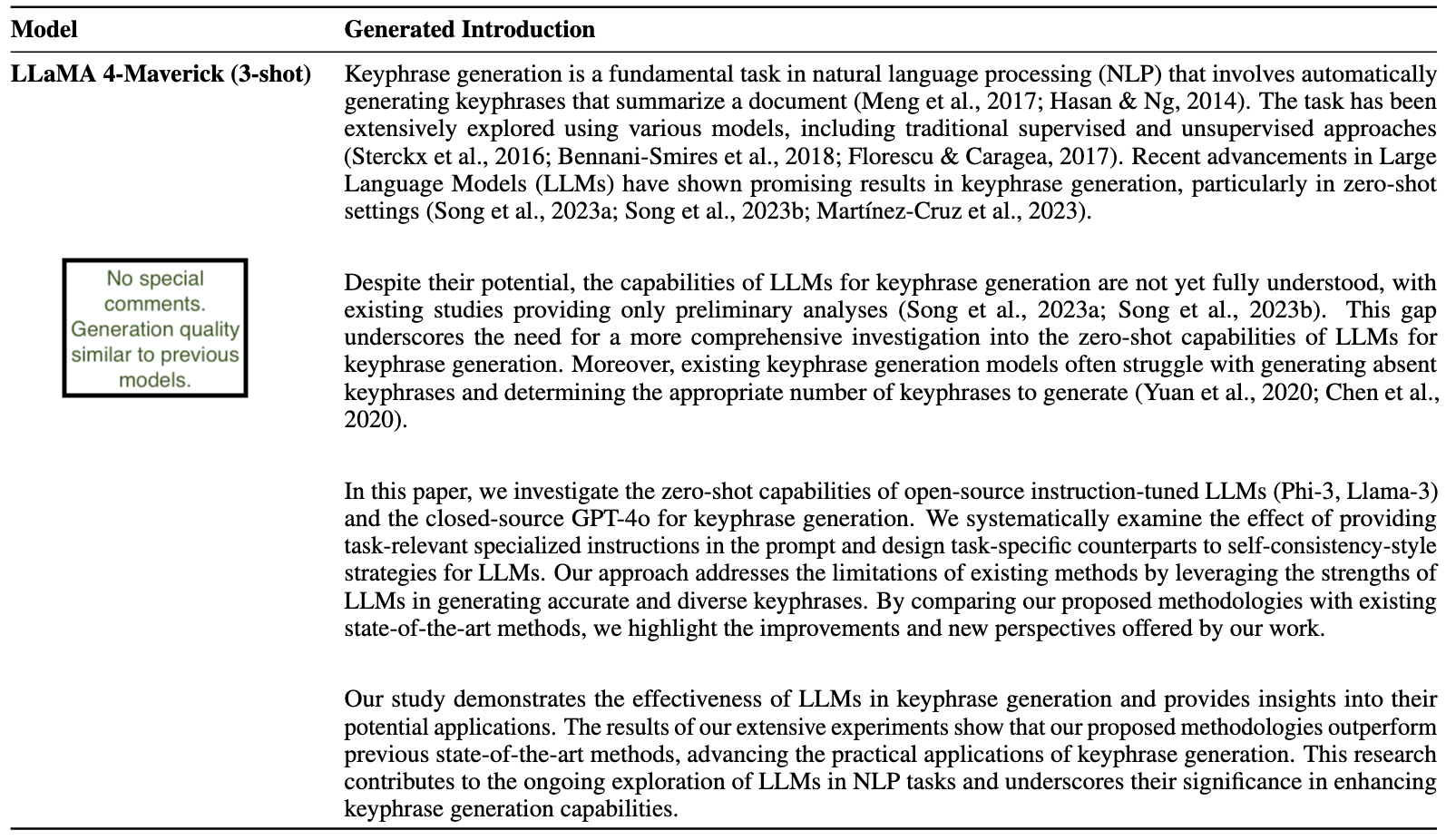}
    \caption{Generated introduction by LLaMA 4-Maverick (3-shot) for ``Zero-Shot Keyphrase Generation: Investigating Specialized Instructions and Multi-Sample Aggregation on Large Language Models.''}
    \label{tab:gen-llama-3shot}
\end{figure*}

\begin{figure*}[t]
  \centering
  \includegraphics[width=\linewidth]{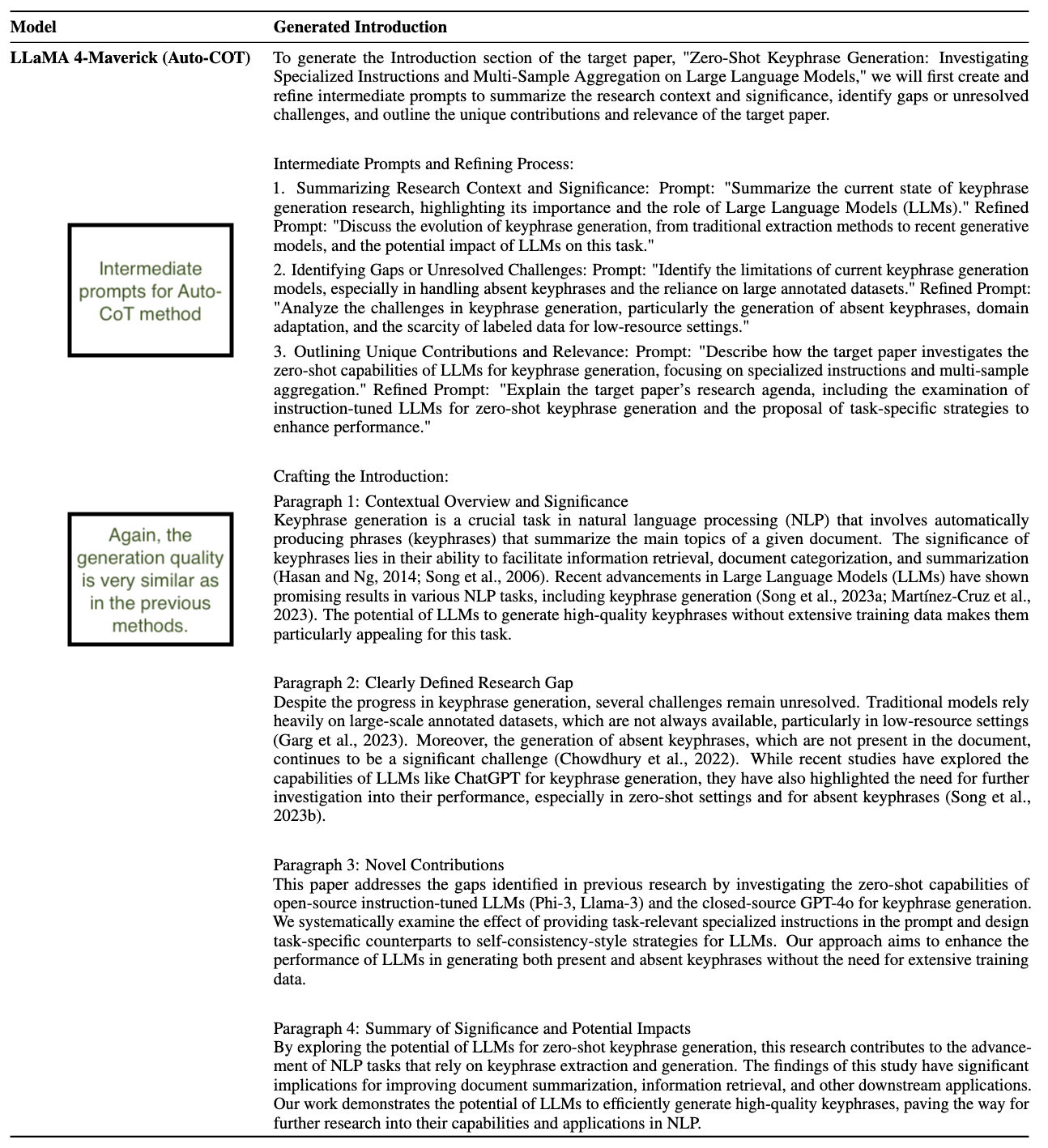}
    \caption{Generated introduction by LLaMA 4-Maverick (Auto-COT) for "Zero-Shot Keyphrase Generation: Investigating Specialized Instructions and Multi-Sample Aggregation on Large Language Models."}
    \label{tab:gen-llama-autocot}
\end{figure*}

% In this section, we have included generated introductions of a sample paper from our test dataset NAACL 2025.
%The generated samples are presented in Figures~\ref{tab:gen-gt}, \ref{tab:gen-gpt4o}, \ref{tab:gen-llama-3shot} and \ref{tab:gen-llama-autocot}.
%In Figures~\ref{tab:gen-gt}, \ref{tab:gen-gpt4o}, \ref{tab:gen-llama-3shot} and \ref{tab:gen-llama-autocot}, we have included a case study for the paper “Zero-Shot Keyphrase Generation: Investigating Specialized Instructions and Multi-Sample Aggregation on Large Language Models.” 
In this section, we do a case study for a NAACL 2025 dataset paper “Zero-Shot Keyphrase Generation: Investigating Specialized Instructions and Multi-Sample Aggregation on Large Language Models”. We have provided the ground truth as well as the generation introductions by four different strategies including LLaMA 4-Maverick (Elaborate, 3-shot, Auto-COT) and GPT-4o (Elaborate) in Figures~\ref{tab:gen-gt}, \ref{tab:gen-gpt4o}, \ref{tab:gen-llama-3shot} and \ref{tab:gen-llama-autocot}.

For the given sample, annotators consistently praised the models, across LLaMA-4 Maverick and GPT-4o, for producing fully faithful introductions that “didn’t introduce any unsupported claims or facts” and maintained a strong narrative flow. They also highlighted that the models captured “non-trivial citations and used them in the right context,” demonstrating strong citation integration even under zero-shot conditions.

Despite this, annotators repeatedly noted that the generated introductions lacked essential conceptual details, particularly the “three research questions,” which were not present in the title, abstract, or related works. This led to lower consistency scores, with annotators clarifying that such omissions were “understandable since the inputs do not give out these details anywhere except the Ground Truth Introduction.” These comments underscore a key limitation: LLMs struggle to reconstruct paper-specific structure when that structure is missing from the input.

Annotators also observed quality differences in background grounding. While LLaMA-4 produced “nice background” sections using related-work abstracts, GPT-4o occasionally drifted into “irrelevant discussion and citations.” A universal weakness noted across all configurations was the Contributions Summary, which annotators described as “looking more like a conclusion” rather than a proper summary of contributions, again reflecting the absence of explicit contribution statements in the prompting material.

Although all models produced strong motivation sections, annotators highlighted two consistent weaknesses: the LLMs failed to generate bullet-style contribution statements—often producing conclusion-like text instead, and they could not recover deeper introduction details such as specific motivations, research questions, or methodological nuances missing from the inputs (title, abstract and related works). A better prompt will likely fix the conclusion issue, but the deeper nuances issue is a limitation of the current setup since Results and other sections are not provided as inputs.

\subsection{Full-Paper-Context Case Study}
\label{app:fullcontext}
To further examine whether richer input beyond Title, Abstract, and Related Work would improve generation, we conducted a focused case study on the same paper \emph{``Zero-Shot Keyphrase Generation: Investigating Specialized Instructions and Multi-Sample Aggregation on Large Language Models.''} We prompted GPT-5.1 with the entire paper, excluding the Introduction, and asked it to regenerate the Introduction. Compared to the restricted-input setting (Title, Abstract, Related Work only), the full-context generation was qualitatively stronger: it explicitly recovered the three research questions, accurately summarized the methodological design, and incorporated empirical findings into a coherent concluding framing — all elements that were missing or weakly developed when only Title, Abstract, and Related Work were provided, as previously noted by annotators. This analysis confirms that richer context can improve Introduction quality. However, conditioning on Methodology and Results would also require post-hoc knowledge typically unavailable at writing time, substantially increase token usage on full-paper inputs (5{,}000–10{,}000 words), and depend on reliable structural extraction across heterogeneous PDF formats. Our restricted-input design therefore represents a deliberate trade-off between input richness and the realism, scalability, and reproducibility required for benchmark-scale evaluation.

\end{document}